\documentclass[10pt,twocolumn,letterpaper]{article}
\PassOptionsToPackage{dvipsnames,table}{xcolor}
\usepackage[final]{cvpr}      

\usepackage[dvipsnames]{xcolor}

\newcommand{\method}[1]{RAR} 
\newcommand{\methodM}[1]{A Unified Framework for Iterative Image Restoration} 

\definecolor{cvprblue}{rgb}{0.21,0.49,0.74}

\usepackage{times}
\usepackage{graphicx}
\usepackage{amsmath}
\usepackage{amssymb}
\usepackage{booktabs}
\usepackage{rotating}
\usepackage{caption}
\usepackage{subcaption}
\usepackage{arydshln}
\usepackage{array}
\usepackage{multirow}
\usepackage{ragged2e}
\usepackage{booktabs}
\usepackage{arydshln}
\usepackage{makecell}

\usepackage[dvipsnames]{xcolor}
\usepackage{times}
\usepackage{array}
\usepackage{ragged2e}
\usepackage{booktabs}
\usepackage{wrapfig}
\usepackage{pifont}
\newcommand{\cmark}{\ding{51}}%
\newcommand{\xmark}{\ding{55}}%
\usepackage{xcolor}

\usepackage[pagebackref,breaklinks,colorlinks,citecolor=blue,linkcolor=blue,bookmarks=false]{hyperref}
\usepackage{makecell}
\usepackage[accsupp]{axessibility}
\usepackage{color, colortbl}
\definecolor{LightCyan}{rgb}{0.88,1,1}

\hypersetup{
    colorlinks=true,
    urlcolor=magenta,
    citecolor=blue,
}

\usepackage[T1]{fontenc}
\usepackage{multirow}
\usepackage{booktabs}
\usepackage{comment}
\usepackage{color}
\usepackage{amsmath}

\usepackage{textcomp}
\usepackage{siunitx}
\usepackage{tabulary}
\usepackage{makecell}
\usepackage{multirow}
\usepackage{booktabs}
\usepackage{hyperref}
\usepackage[ruled,vlined]{algorithm2e}

\makeatletter
\newcommand\figref{Figure~\ref}
\newcommand{\tabref}[1]{Table~\ref{#1}}
\newcolumntype{P}[1]{>{\centering\arraybackslash}p{#1}}
\newcolumntype{M}[1]{>{\centering\arraybackslash}m{#1}}
\let\ts@includegraphics\includegraphics

\usepackage{float}
\usepackage{lipsum}   
\usepackage{rotating} 
\usepackage{stfloats} 
\usepackage[export]{adjustbox} 


\def\paperID{39236} 
\def\confName{CVPR}
\def\confYear{2026}

\begin{document}
\title{Restore, Assess, Repeat: A Unified Framework for Iterative Image Restoration}
\vspace{-20mm}\author{
    I-Hsiang Chen\textsuperscript{1,2,*}
    \quad Isma Hadji\textsuperscript{1}
    \quad Enrique Sanchez\textsuperscript{1}
    \quad Adrian Bulat\textsuperscript{1,3}
    \quad Sy-Yen Kuo\textsuperscript{2,4}\\
    \quad Radu Timofte\textsuperscript{5}
    \quad Georgios Tzimiropoulos\textsuperscript{1,6}
    \quad Brais Martinez\textsuperscript{1}
    \\\\
    \hspace{-8mm}\textsuperscript{1}Samsung AI Center Cambridge\quad \textsuperscript{2}National Taiwan University\quad \textsuperscript{3}Technical University of Iasi \\ \textsuperscript{4} Chang Gung University \quad \textsuperscript{5}University of Wurzburg \quad \textsuperscript{6}Queen Mary University of London
}

\maketitle


\begin{abstract}
Image restoration aims to recover high quality images from inputs degraded by various factors, such as adverse weather, blur, or low light. While recent studies have shown remarkable progress across individual or unified restoration tasks, they still suffer from limited generalization and inefficiency when handling unknown or composite degradations. To address these limitations, we propose \method{}, a Restore, Assess and Repeat process, that integrates Image Quality Assessment (IQA) and Image Restoration (IR) into a unified framework to iteratively and efficiently achieve high quality image restoration. Specifically, we introduce a restoration process that operates entirely in the latent domain to jointly perform degradation identification, image restoration, and quality verification. The resulting model is fully trainable end-to-end and allows for an all-in-one assess and restore approach that dynamically adapts the restoration process. Also, the tight integration of IQA and IR into a unified model minimizes the latency and information loss that typically arises from keeping the two modules disjoint, (e.g. during image and/or text decoding). Extensive experiments show that our approach offers consistent improvements under single, unknown and composite degradations, thereby establishing a new state-of-the-art. 
\end{abstract}

\newcommand\blfootnote[1]{%
\begingroup
\renewcommand\thefootnote{}\footnote{#1}%
\addtocounter{footnote}{-1}%
\endgroup
}
\blfootnote{Project: \url{https://restore-assess-repeat.github.io/}}
\blfootnote{* Work conducted during I-Hsiang\textquotesingle s internship at Samsung AI Center, Cambridge, UK.}

\setlength{\parskip}{0em}
\vspace{-20pt}
\section{Introduction}\label{sec:introduction}

Image Restoration is a challenging task that aims at recovering a clean image from one or several degradations such as blur, haze, rain, or lighting. Most early works focused on tackling one kind of degradation only~\cite{chihaoui2024self, ren2023multiscale, zamir2021multi, yi2023diff}. However, in real world settings, degradations are complex and rarely known a priori. In here, we focus on image restoration under unknown composite degradations, so images can present one or more unknown degradations at a time.

 \begin{figure}[t]
   \begin{center}
    \includegraphics[width=0.99\linewidth]{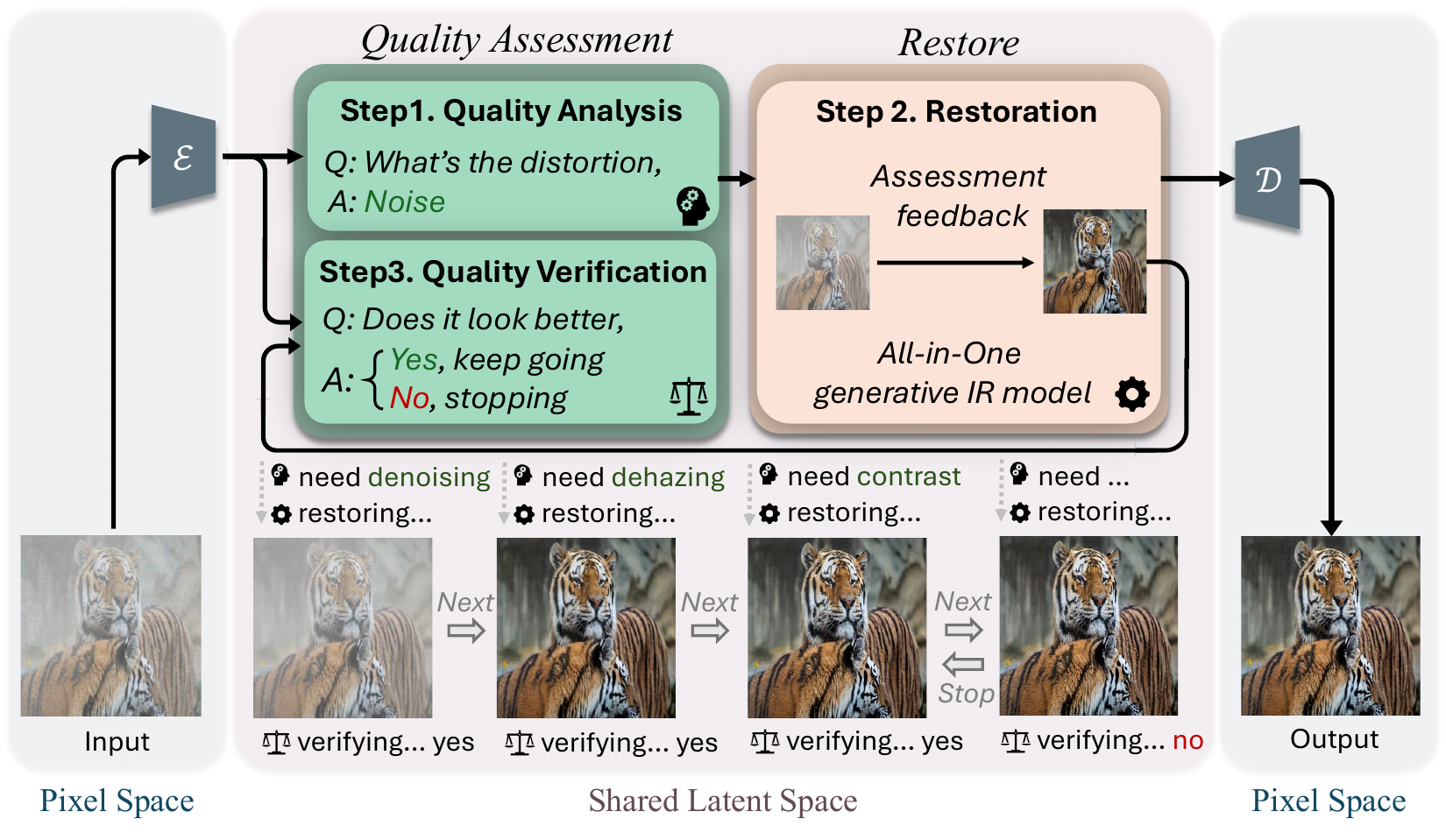}
    \caption{\textbf{Overview of \method{}}. Given an input image with unknown composite degradations, \method{} operates entirely in latent by iteratively assessing current image quality, restoring the image based on the assessment, and verifying the quality improvement.}
    \label{fig:main_fig}
    \end{center}
    \vspace{-20pt}
 \end{figure}

Recent methods tackling composite degradations can be broadly categorized into two paradigms. The first, \textit{all-in-one} models~\cite{zheng2024selective, lin2023diffbir, autodir, restorevar} utilize a single, unified generative model for restoration.
The second, \textit{agentic} models~\cite{agenticir,rlrestore,restoreagent}, employ an agent that iteratively selects and applies tools from a suite of specialized, single-degradation restorers. Tool selection is guided by an IQA step, which identifies the main degradation. While agentic models allow for progressive restoration, they are cumbersome and slow. Conversely, all-in-one models offer a more efficient and unified
alternative, but currently lag in performance. 

AutoDIR \cite{autodir}, one of the best-performing all-in-one models, also uses an assessment step in line with agentic models. However, it relies on CLIP~\cite{clip}, which treats IQA as n-way classifier over a predefined set of degradation labels. 
This limits the generalization capability of the restoration model. Furthermore, zero-shot CLIP performs poorly for IQA. 
Thus, supervised finetuning is required to make it focus on degradations rather than semantics. 
Instead, recent agentic models (e.g AgenticIR~\cite{agenticir}) rely on a description-based IQA model~\cite{depictqa}, which provides a rich free-form assessment of the image and covers multiple assessment tasks. 
However, given the lack of integration with the restoration step, 
even when the IQA explicitly identifies image degradations, final performance is still limited by the availability of a corresponding restoration tool and proper execution.

In this work, we take the best of both worlds by integrating a more descriptive IQA model, capable of assessing image quality through free form text, into an all-in-one generative IR model. Importantly, we propose a deep integration of the two modules such that they share a common latent space, thereby yielding a single end-to-end trainable model that is capable of multiple rounds of degradation identification, restoration and quality verification in latent space.

To achieve this, we start from an IQA model that can describe the degradations in an image using free form text but enforce the IQA model to operate in the same latent space as the IR module. For the IR we use a generative model, which we condition on the feedback obtained from the IQA model directly using its output logits rather than the decoded text, thereby tightly combining the assessment and restoration steps. The resulting model is fully trainable in an end-to-end fashion to dynamically adapt the IR on the IQA feedback. At inference, the IQA model is also used to compare images before and after each restoration step to decide that (a) the restoration process is successful or (b) repeat the process with a new conditioning based on trailing issues in the restored image.
The entire process is illustrated in Fig.~\ref{fig:main_fig}. In summary, our contributions are threefold:
\begin{itemize}
\item We propose to combine free-form IQA with an all-in-one restoration model.
\item We tightly integrate the IQA and restoration into a single end-to-end trainable model.
\item We enable multiple assessment, restoration, validation rounds, all operating in latent space, thereby improving both training process and inference performance.
\end{itemize}
Together, these contribution yield state-of-the-art performance on unknown, composite and single degradation settings. For example, in the challenging composite degradations setup, our method achieves improvements ranging from +2.71 dB (PSNR) to +72.4 (MANIQA), while being 11.27× faster than the state-of-the-art.

\section{Related Work}
\label{sec:related_work}
Image restoration is a challenging inverse imaging task. Early works focused on single-degradation settings such as denoising~\cite{zamir2021multi,chihaoui2024self}, deblurring~\cite{nah2017deep, ren2023multiscale}, dehazing~\cite{dong2014learning, chen2019pms,zamir2021multi}, deraining~\cite{wang2020model,yang2017deep} or low-light enhancement~\cite{yi2023diff,zhang2023unified}. 
To better align with real-world conditions, recent works address multiple degradations, which is our main focus in this work. We categorize methods into two main tacks: all-in-one models and agentic models.

\noindent\textbf{All-in-one models} rely on a single model to tackle multiple degradations. These methods are either non-generative \cite{chen2022simple, zamir2021multi, liang2021swinir, zamir2022restormer, li2020all, valanarasu2022transweather, potlapalli2024promptir}, or rely on a generative model \cite{ozdenizci2023restoring, zhu2023denoising, xia2023diffir,zheng2024selective, lin2023diffbir, chen2025unirestore, restorevar}. Most such methods use known degradation priors or a classifier to recognize the degradation category present in the image to guide the restoration model. Therefore, these methods are constrained to handle a limited set of degradations. While we also rely on generative priors to perform restoration, we do not assume knowledge of the degradation. Instead, we enlist an IQA model to condition the restoration process. Most closely related work to ours is AutoDIR \cite{autodir}, which attempts to handle unknown degradations also using an IQA model. However, they rely on CLIP, which by default performs poorly for degradation identification, thus needing finetuning to operate on a predefined closed set of degradations. Therefore, it inherits the shortcoming of prior methods. Instead, we start from DepictQA~\cite{depictqa}, a VLM-based IQA, which not only can describe image quality with free-form text, but also can compare pairs of images. We then adapt it to operate in the latent space of a restoration module, to yield an end-to-end trainable model, which enables an iterative assess and restore process.
\\\\
\noindent \textbf{Agentic Models.} instead consider a large set of models (tools) that specialize on different single degradations \cite{rlrestore,restoreagent,agenticir}. They then either rely on reinforcement learning to define a policy that select appropriate tools \cite{rlrestore}, or finetune LLM/VLM models to iteratively select and apply the different tools considered \cite{restoreagent, agenticir}. We also use a VLM-based IQA, however, agentic model do not fully integrate the IQA and IR modules into a unified model. Hence, they are typically slow and need more steps to guide the restoration. For example, AgenticIR \cite{agenticir} starts from a similar IQA to ours. However, given the disconnect between the IQA and IR tools, it requires the restoration models to first decode their output into actual images, which are then encoded in the latent space of the IQA to get an assessment, which is then passed to a powerful LLM \cite{openai2023gpt4} to devise a restoration plan that is tested in a trial-and-error manner, thereby making the entire process extremely inefficient. In addition, even when the IQA describes the correct degradations, the restoration process is limited by the availability of a restoration tool in the predefined set available to the agent. Our work takes the best of both worlds using a strong VLM-based IQA and fully integrating it into a strong all-in-one generative IR model, attaining superior performance and better efficiency.
\vspace{-15pt}

\section{Proposed Method}\label{sec:method}

\begin{figure*}[t]
    \centering \includegraphics[width=\linewidth]{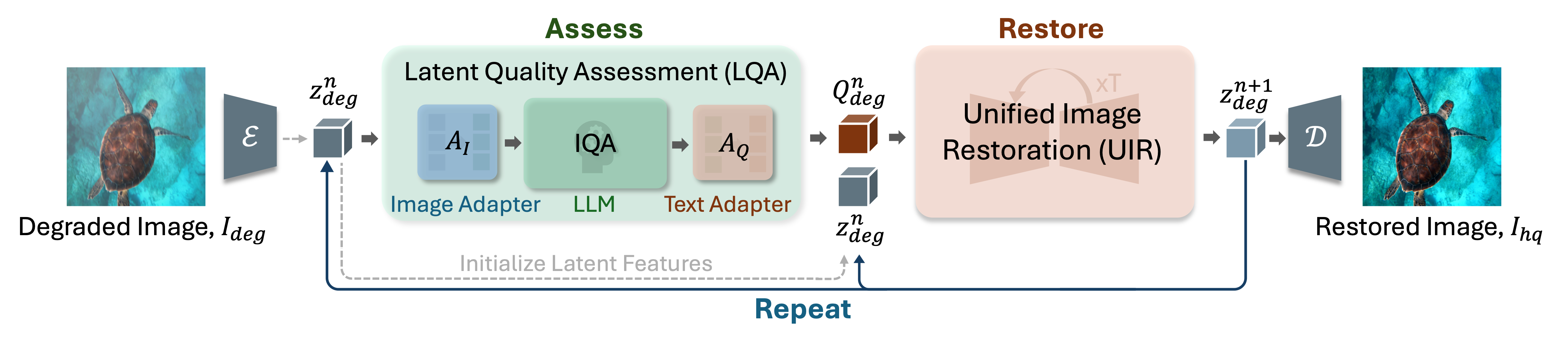}
    \caption{To enable our \method{} process, we define an LQA that fully integrates the input and out of the IQA into the latent space of the restoration module, using adapters $\mathcal{A}_I$ and $\mathcal{A}_Q$, respectively. Then, we instore a feedback loop between the two modules to iteratively recover best image quality. The dashed-line is only used for the very first iteration to feed $\mathbf{z}^0_{deg}$ to the restoration model.}
    \label{fig:overall-method}
\end{figure*}

\subsection{Preliminaries}
\label{ssec:strong_baseline}

Given a degraded image, $I_{deg} \in \mathbb{R}^{H\times W \times C}$, our goal is to automatically identify image degradations using an IQA module, and to progressively restore the image into a high quality version, $I_{hq} \in \mathbb{R}^{H\times W \times C}$, by relying on the IQA's assessment for conditioning. 
To this end, we consider a framework consisting of two main modules; namely, (1) an Image Quality Assessment (IQA) module that can describe image quality in free-form text and (2) an Image Restoration (IR) module that can be conditioned on such text.

We have the following key requirements for the IQA module: (i) we want it to generalize rather than focus on a closed set of degradation labels, (ii) handle composite degradations and (iii) provide rich conditioning information for the restoration module. To this end, we use the recently introduced DepictQA \cite{depictqa}, which we further finetune for our purposes. Note that previous work built upon the same IQA model to activate different restoration agents~\cite{agenticir}.
For the IR module, we can take advantage of the strong priors learned by any state-of-the-art generative model, such as a standard diffusion model \cite{rombach2022high}. We term this component as Unified Image Restoration (UIR) model.

A naive implementation would perform IQA on the degraded image to obtain a free-form description of the degradations present, and condition the restoration process on the output of the IQA by using the text-conditioning branch of diffusion models.
More formally:
\begin{eqnarray}
Q_{deg} & = & IQA(I_{deg})\\
I_{hr} & = &Restore(I_{deg}|Q_{deg})
\end{eqnarray}

However, we have two main extra desiderata in respect to this naive integration: (a) we would like the IQA and the restoration modules to be tightly integrated to better exploit their synergetic relationship, both at inference and training time. (b) we would like an iterative process where the IQA assessment is updated throughout the restoration process to better adapt to the current restoration status. To this end, we introduce Latent Quality Assessment (LQA) in Sec.~\ref{sec:lqa}, which results in an end-to-end trainable unified model. Then we describe our iterative restoration process and adaptive stopping criteria, which rely on intermediate quality assessments in Sec.~\ref{sec:rar}.

\subsection{Latent Quality Assessment}
\label{sec:lqa}
The naive integration in the previous section treats the IQA and UIR as independent models. In this section, we unify them into a single end-to-end trainable model. This requires two main changes: (i) aligning the input latent spaces of the IQA and the restoration models, (ii) aligning the output IQA latents and the restoration conditioning latents. 

\noindent \textbf{Latent space alignment:} In the naive integration, the restoration model first projects the degraded image $I_{deg}$ into latent space using an auto-encoder $\mathcal{E}_{restore}$, obtaining $\mathbf{z}_{deg}=\mathcal{E}_{restore}(I_{deg})$. Instead, the IQA operates from pixel space and uses a different auto-encoder $\mathcal{E}_{IQA}$. We instead modify the input to the IQA to obtain a Latent Quality Assessment (LQA) model,
\begin{equation}
    LQA(\mathbf{z}_{deg}) = IQA(\mathcal{A}_I(\mathbf{z}_{deg}))
\end{equation}
where $\mathcal{A}_I$ is an adapter module that we train to bridge the gap between the two representations. We follow a two-stage finetuning process. First, we only finetune the image adapter, $\mathcal{A}_I$, while keeping the IQA weights frozen. We then unfreeze all layers to further finetune the full LQA. 

\noindent \textbf{Conditioning alignment:} The LQA module described above produces an assessment text form. This is then encoded by the text-conditioning branch of the diffusion model to produce the conditioning embeddings. The decoding step is undesirable because it is non-differentiable and also loses information compared to the IQA output latents.
Furthermore, conditioning based on the IQA text output relies on the parameter and latency-heavy text-conditioning branch of the restoration model. 
We propose instead to directly align 
the latent output of the IQA, $\tilde{Q}_{deg}=LQA(\mathbf{z}_{deg})$, to the output embeddings of the text-conditioning branch of the restoration model, $\mathcal{T}$. Once again, we use an adapter $A_Q$ to align the two representations such that 
\begin{equation}
    Q_{deg} = \mathcal{T}(\mathcal{A}_Q(\tilde{Q}_{deg}))
\end{equation}
We train in two stages, first finetuning the text adapter, $\mathcal{A}_Q$, before unfreezing the UIR weights in the second stage. Notably, with this step, we have removed the need for decoding between the assessment and restoration stage, thereby yielding an end-to-end trainable model. Furthermore, the text-conditioning branch on the restoration model can now be dropped entirely, saving both the latency and the parameter count. The details architecture and training are provided in the supplemental material. 

\subsection{Restore, Assess, Repeat}\label{sec:rar}
\begin{figure}[t]
    \centering \includegraphics[width=\columnwidth]{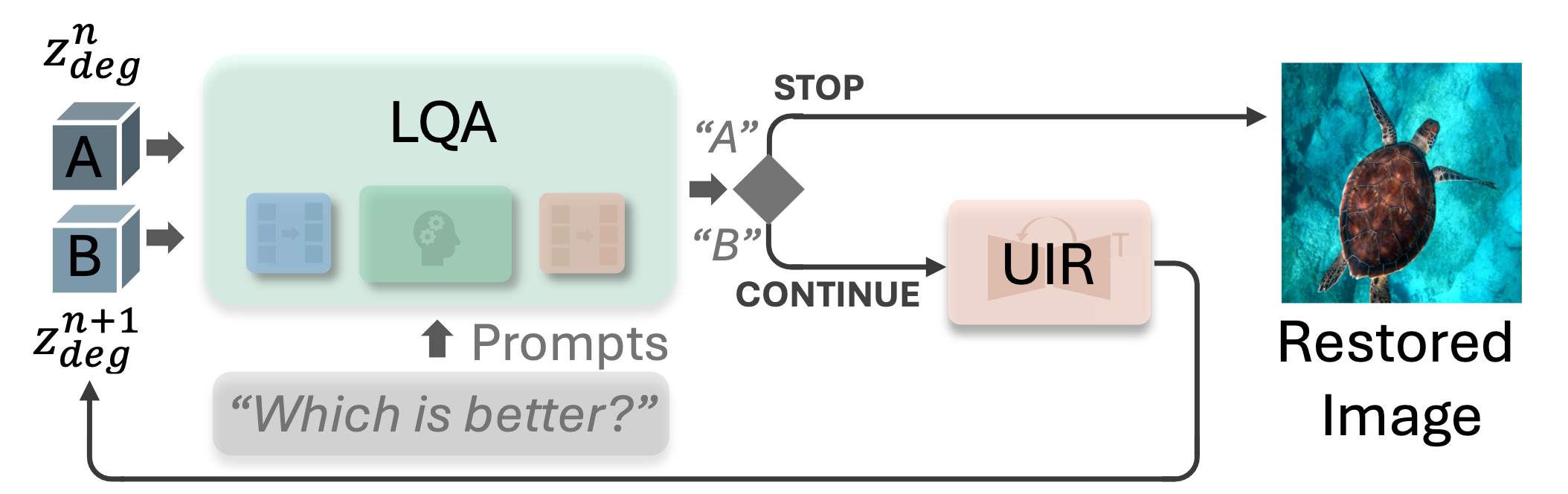}
    \caption{We illustrate our quality verification step. This is used at inference time to define a stopping criterion of the \method{} process.}
    \label{fig:stopping}
\end{figure}
Up until now, our approach has been agnostic of the restorer module employed, as long as it can be conditioned on the output of the IQA. 
However, now we want to dynamically update the conditioning signal provided by the LQA. This requires re-assessing the quality using intermediate latents, $\mathbf{z}^n_{deg}$, as input (e.g. output of denoising/flow step $n$).
If we adopt the diffusion-based denoising paradigm, intermediate representations will contain added noise, leading to inaccurate readings of the LQA. We thus require a model that does not involve a noising process with every update to the input condition.
We therefore rely on Stable Diffusion 3.5 (SD3.5) \cite{SD3} as it uses Flow Matching (FM). 

While FM learns a direct mapping between input and output distributions, it is common for image editing to learn a mapping from noise, $\mathcal{N}\sim(0,I)$, to the target distribution, and include the degraded image as part of the conditioning context. 
Instead, we remove the noise term and learn a direct mapping from the degraded image distribution, $\rho_{deg}$, to the high quality image distribution, $\rho_{hq}$, and remove the degraded image from the conditioning context.

More specifically, we define a mapping between the degraded input, $\mathbf{z}_{deg} \sim \rho_{deg}$, and the target output, $\mathbf{z}_{hq} \sim \rho_{hq}$, where $\mathbf{z}$ is the representation of images in the latent space of the UIR model. A model $v_\theta$, is trained to predict the velocity vector between the two distribution such that
\begin{equation}
   \mathcal{L}_v = \mathbb{E}_{\mathbf{z}_{deg},\mathbf{z}_{hq},Q_{deg},t} || v_\theta(\mathbf{z}_t,Q_{deg},t) - (\mathbf{z}_{hq}-\mathbf{z}_{deg})||^2 \label{eq:luir}
\end{equation}
where $\mathbf{z}_t$ is a linear interpolation between the two distributions, defined as $\mathbf{z}_t = (1 - t) \mathbf{z}_{deg} + t \mathbf{z}_{hq}$, and $v = v_\theta(\mathbf{z}_t, Q_{deg},t)$ is the vector pointing from $\mathbf{z}_{deg}$ to $\mathbf{z}_{hq}$.

In Table.~\ref{tab:abla_3} we show, quantitatively, the importance of this formulation, which allows us to iteratively update the input, $\mathbf{z}_{deg}$, without being corrupted by added noise.

\noindent \textbf{Iterative assessment:} We now take full advantage of the multi-step nature of the adopted flow-based model to iteratively assess the restoration progress by adding a feedback loop between the LQA and the UIR modules. In particular, the UIR model starts the restoration process by taking a degraded image latent $\mathbf{z}^0_{deg}$ and its corresponding LQA assessment $Q^0_{deg}$ as input. The input conditioning is then updated dynamically after $n$ iterations. Specifically, the training objective in \eqref{eq:luir} is updated such that
\begin{equation}
    \mathcal{L}_v = \mathbb{E}_{\mathbf{z}^n_{deg},\mathbf{z}_{hq},Q^n_{deg},t} || v_\theta(\mathbf{z}^n_t,Q^n_{deg},t) - (\mathbf{z}_{hq}-\mathbf{z}^n_{deg})||^2
\end{equation}

\noindent with $\mathbf{z}^n_t = (1 - t) \mathbf{z}^n_{deg} + t \mathbf{z}_{hq}$, and $v^n = v_\theta(\mathbf{z}^n_t, Q^n_{deg},t)$. The predicted velocity is used periodically to update the input such that $\mathbf{z}^{n+1}_{deg} = \mathbf{z}^{n}_{deg} + v^n$. Similarly, the conditioning term is updated using the integrated LQA module such that $Q^{n+1}_{deg} = LQA(\mathbf{z}^{n+1}_{deg})$. Thanks to our formulation with direct mapping from degraded distribution to target distribution, the LQA can provide meaningful feedback without being affected by noise, as shown in Tab.~\ref{tab:abla_3}. 

Fig.~\ref{fig:main_fig} exemplifies this process, showing an input image (leftmost side) that contains multiple degradations. The restoration model removes the noise first. Rather than insisting on denoising, the resulting latent is then fed back to the LQA, that now identifies haze as the remaining degradation. The restorer takes this updated feedback and removes haze in the next iteration. 

During training, our \method{} process is seamlessly integrated into standard flow-matching training, where the model is typically trained with a predefined number of timesteps and the LQA can be called for any random timestep. At inference time, we perform an assessment every $T$ steps and stop the iterations upon reaching a stopping criterion, as described next. The fully integrated model is illustrated in Fig.~\ref{fig:overall-method}.

\noindent \textbf{RAR stopping criterion}: To devise a stopping criterion for the proposed \method{} process, we again use the LQA model. Given that our LQA model is based on DepictQA \cite{depictqa}, which can compare pairs of images, we re-enlist it to define a stopping criterion. Specifically, after every $T$ steps through our model, we use the LQA to compare the quality of the images before and after restoration. As illustrated in Fig.~\ref{fig:stopping}, the LQA takes latents $\mathbf{z}^n_{deg}$ and $\mathbf{z}^{n+T}_{deg}$ as input and is tasked to compare their quality and yield a binary decision to either \texttt{CONTINUE} if the quality of the new latent is better than the previous, or \texttt{STOP}, otherwise. In the latter case, we use $\mathbf{z}^n_{deg}$ as our final model prediction.
\section{Experiments}
\label{sec:exp}

We now thoroughly evaluate the effectiveness of the proposed \method{} process. We first define our experimental setup in Sec.~\ref{sec:setup}. We then provide extensive quantitative and qualitative comparisons to state-of-the-art in Sec.~\ref{sec:compare}. 
Finally, we support each component in our framework with a thorough ablation study in Sec.~\ref{sec:ablation}. We provide implementation details in the supplementary material.

\subsection{Experimental Setup}\label{sec:setup}

\noindent \textbf{Training Dataset.} We consider the same restoration tasks used in previous work~\cite{autodir, luo2023controlling}; namely, denoising, deblurring, super-resolution, low-light enhancement, dehazing, deraining, and deraindrop. For super-resolution, we use DIV2K~\cite{agustsson2017ntire} and Flickr2K~\cite{Lim_2017_CVPR_Workshops} training sets, following RealESRGAN \cite{wang2021real} for degraded image generation.  
For denoising, we use SIDD~\cite{abdelhamed2018high} and a synthetic Gaussian and Poisson noise dataset with DIV2K~\cite{agustsson2017ntire} and Flickr2K~\cite{Lim_2017_CVPR_Workshops}. 
We also use the training set from GoPro~\cite{nah2017deep}, LOL~\cite{wei2018deep}, RESIDE~\cite{li2017reside}, Rain200L~\cite{yang2017deep}, and Raindrop~\cite{qian2018attentive} for deblurring, low-light enhancement, dehazing and  deraining. 
For the Latent Quality Assessment, we followed the DepictQA~\cite{depictqa} pipeline and extended its training corpus to include the aforementioned restoration datasets.
%

\noindent \textbf{Testing Dataset.} During inference, we categorize image restoration into three tasks: composite degradations, single degradation and unknown degradations. 
For the \textit{composite degradations}, we followed the AgenticIR~\cite{agenticir} setting, which constructs 16 combinations of mixed degradations based on the MiO100~\cite{kong2024towards} dataset. These combinations are divided into three subsets (A,B, and C), where subsets A and B involve two degradations and subset C involves three, simulating progressively complex degradation scenarios. 
For the \textit{single degradation}, we followed the AutoDIR~\cite{autodir} setting, where single degradation restoration is evaluated on standard benchmarks, including SIDD~\cite{abdelhamed2018high}, Kodak24~\cite{franzen1999kodak}, DIV2K~\cite{agustsson2017ntire}, GoPro~\cite{nah2017deep}, LOL~\cite{wei2018deep}, SOTS-Outdoor~\cite{li2018benchmarking}, Rain100~\cite{yang2017deep}, and Raindrop~\cite{qian2018attentive} dataset. 
For the \textit{unknown degradations}, we used UDC~\cite{zhou2021image} and EUVP~\cite{islam2020fast} datasets to assess generalization to unseen degradation types.

\noindent \textbf{Baselines:} 
To evaluate the effectiveness of our method, we compare it with multiple All-in-One restoration approaches, including AirNet~\cite{AirNet}, PromptIR~\cite{potlapalli2024promptir}, MioIR~\cite{kong2024towards}, DA-CLIP~\cite{luo2023controlling}, InstructIR~\cite{conde2024instructir}, AdaIR~\cite{cui2025adair}, DARCOT~\cite{tang2025degradation} and AutoDIR~\cite{autodir}. For completeness, we also compare to AgenticIR \cite{agenticir}, one of the most recent agent-based restoration frameworks. 

\noindent \textbf{Metrics:} 
We consider both fidelity (e.g., PSNR, SSIM) and perceptual metrics (e.g., LPIPS, MANIQA, CLIP-IQA, MUSIQ, NIQE and NIMA), providing balanced assessment between restoration accuracy and perceptual performance. 

\begin{figure}[t]
    \scriptsize
    \centering
    \begin{minipage}[t]{0.33\linewidth}
    \vspace{-\fboxsep}
    \includegraphics[width=\linewidth]{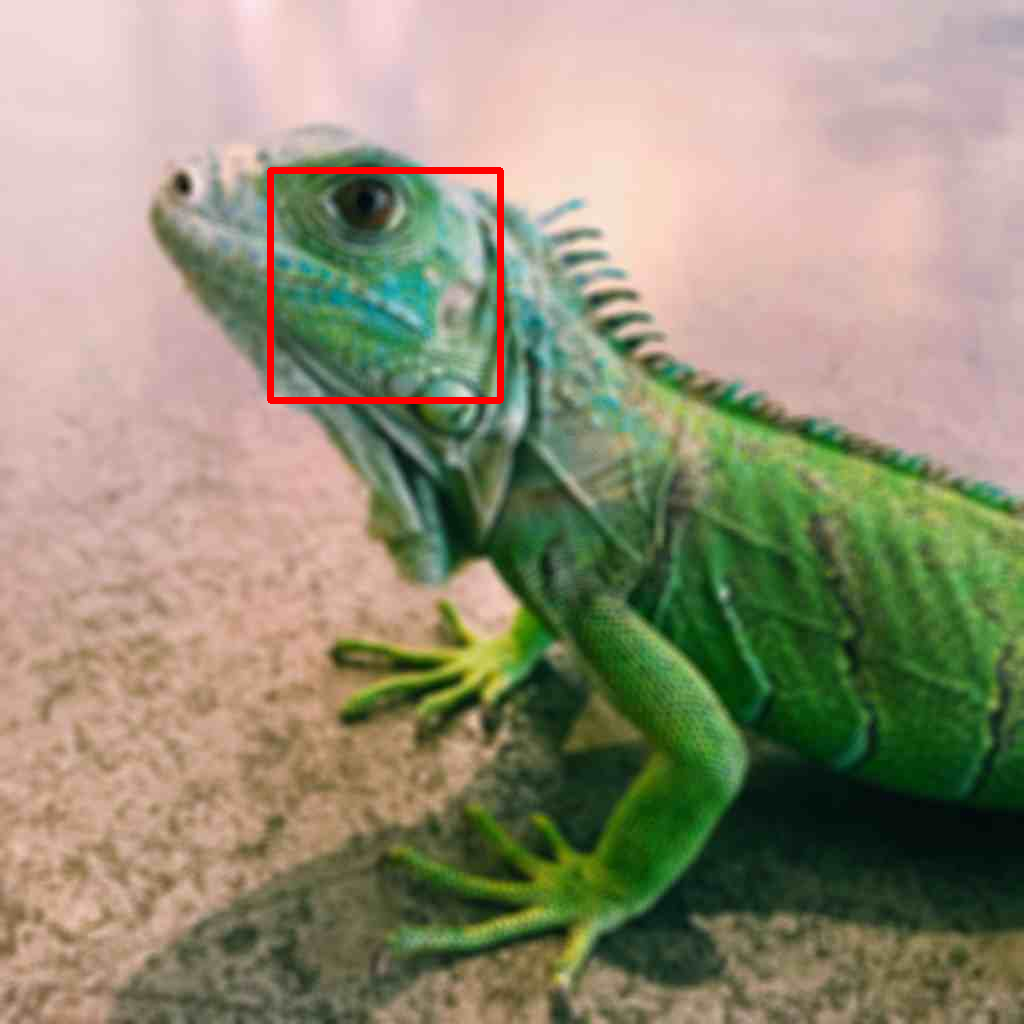}
    \\
    Input image with defocus blur and JPEG
    \end{minipage}
    \hspace{1pt}
    \begin{minipage}[t]{0.63\linewidth}
    \vspace{-\fboxsep}
    \begin{minipage}{0.24\linewidth}
    \includegraphics[width=\linewidth]{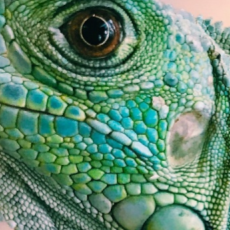}
    \end{minipage}
    \hfill
    \begin{minipage}{0.24\linewidth}
    \includegraphics[width=\linewidth]{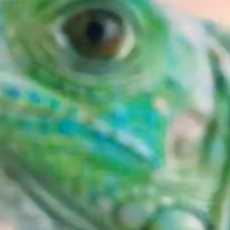}
    \end{minipage}
    \hfill
    \begin{minipage}{0.24\linewidth}
    \includegraphics[width=\linewidth]{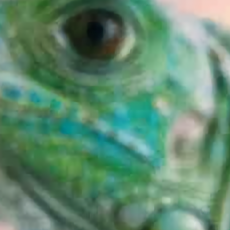}
    \end{minipage}
    \hfill
    \begin{minipage}{0.24\linewidth}
    \includegraphics[width=\linewidth]{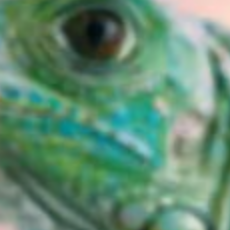}
    \end{minipage}
    \\
    \begin{minipage}{0.24\linewidth}
        \centering
        GT
    \end{minipage}
    \hfill
    \begin{minipage}{0.24\linewidth}
        \centering
        PromptIR
    \end{minipage}
    \hfill
    \begin{minipage}{0.24\linewidth}
        \centering
        MiOIR
    \end{minipage}
    \hfill
    \begin{minipage}{0.24\linewidth}
        \centering
        DA-CLIP
    \end{minipage}  
    \\
    \begin{minipage}{0.24\linewidth}
    \includegraphics[width=\linewidth]{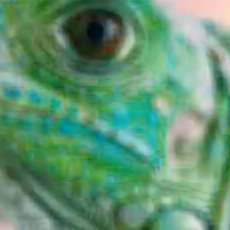}
    \end{minipage}
    \hfill
    \begin{minipage}{0.24\linewidth}
    \includegraphics[width=\linewidth]{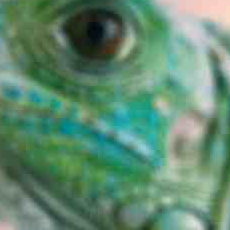}
    \end{minipage}
    \hfill
    \begin{minipage}{0.24\linewidth}
    \includegraphics[width=\linewidth]{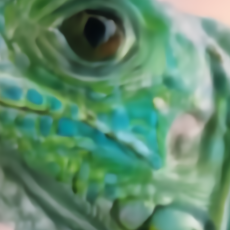}
    \end{minipage}
    \hfill
    \begin{minipage}{0.24\linewidth}
    \includegraphics[width=\linewidth]{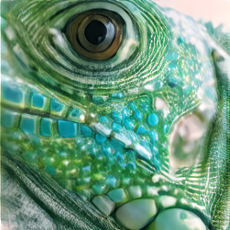}
    \end{minipage}
    \\
    \begin{minipage}{0.24\linewidth}
        \centering
        InstructIR
    \end{minipage}
    \hfill
    \begin{minipage}{0.24\linewidth}
        \centering
        AutoDIR
    \end{minipage}
    \hfill
    \begin{minipage}{0.24\linewidth}
        \centering
        AgenticIR
    \end{minipage}
    \hfill
    \begin{minipage}{0.24\linewidth}
        \centering
        \method{}
    \end{minipage}  
    \\
    \end{minipage}

    \vspace{10pt}

\begin{minipage}[t]{0.33\linewidth}
    \vspace{-\fboxsep}
    \includegraphics[width=\linewidth]{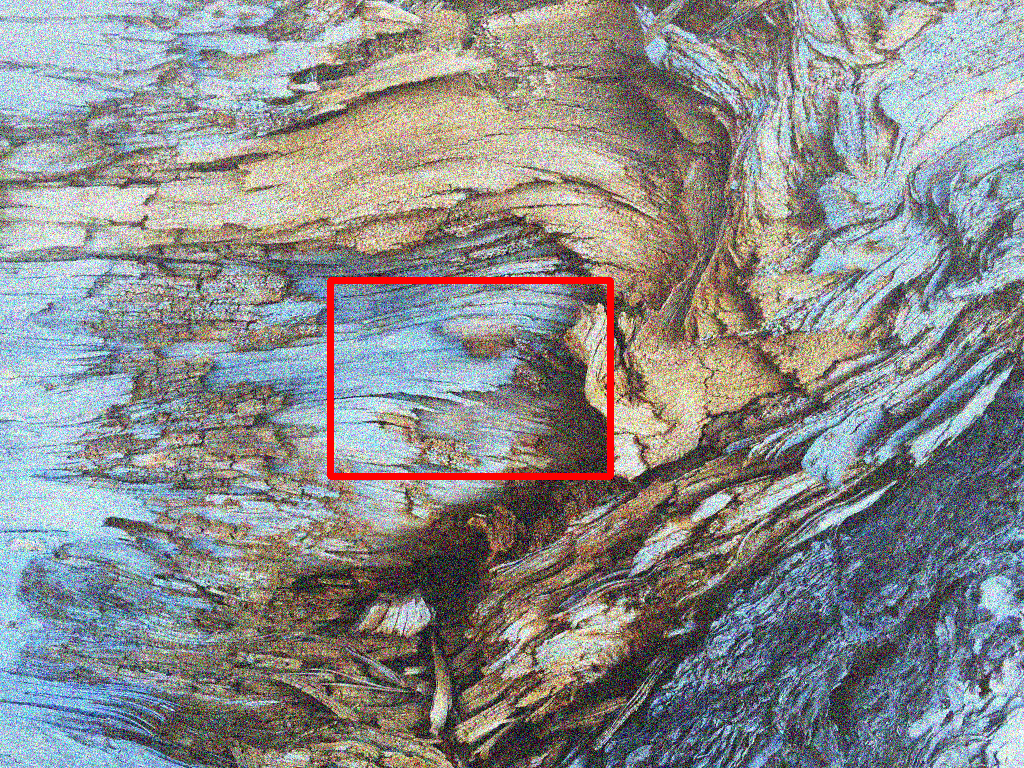}
    \\
    Input image with haze and noise
    \end{minipage}
    \hspace{1pt}
    \begin{minipage}[t]{0.63\linewidth}
    \vspace{-\fboxsep}
    \begin{minipage}{0.24\linewidth}
    \includegraphics[width=\linewidth]{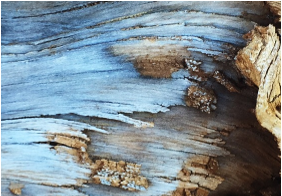}
    \end{minipage}
    \hfill
    \begin{minipage}{0.24\linewidth}
    \includegraphics[width=\linewidth]{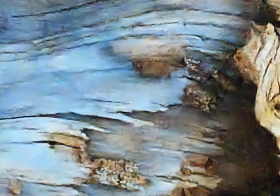}
    \end{minipage}
    \hfill
    \begin{minipage}{0.24\linewidth}
    \includegraphics[width=\linewidth]{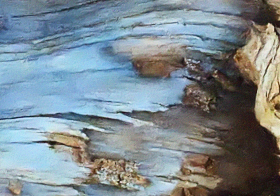}
    \end{minipage}
    \hfill
    \begin{minipage}{0.24\linewidth}
    \includegraphics[width=\linewidth]{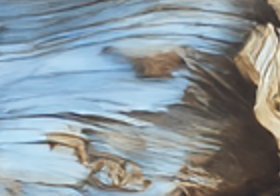}
    \end{minipage}
    \\
    \begin{minipage}{0.24\linewidth}
        \centering
        GT
    \end{minipage}
    \hfill
    \begin{minipage}{0.24\linewidth}
        \centering
        PromptIR
    \end{minipage}
    \hfill
    \begin{minipage}{0.24\linewidth}
        \centering
        MiOIR
    \end{minipage}
    \hfill
    \begin{minipage}{0.24\linewidth}
        \centering
        DA-CLIP
    \end{minipage}  
    \\
    \begin{minipage}{0.24\linewidth}
    \includegraphics[width=\linewidth]{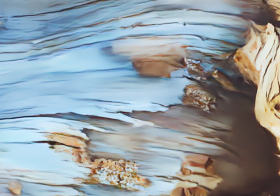}
    \end{minipage}
    \hfill
    \begin{minipage}{0.24\linewidth}
    \includegraphics[width=\linewidth]{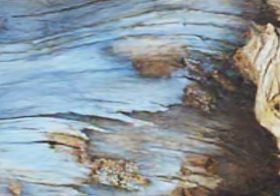}
    \end{minipage}
    \hfill
    \begin{minipage}{0.24\linewidth}
    \includegraphics[width=\linewidth]{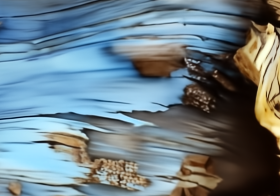}
    \end{minipage}
    \hfill
    \begin{minipage}{0.24\linewidth}
    \includegraphics[width=\linewidth]{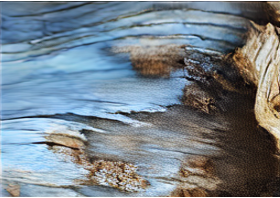}
    \end{minipage}
    \\
    \begin{minipage}{0.24\linewidth}
        \centering
        InstructIR
    \end{minipage}
    \hfill
    \begin{minipage}{0.24\linewidth}
        \centering
        AutoDIR
    \end{minipage}
    \hfill
    \begin{minipage}{0.24\linewidth}
        \centering
        AgenticIR
    \end{minipage}
    \hfill
    \begin{minipage}{0.24\linewidth}
        \centering
        \method{}
    \end{minipage}  
    \\
    \end{minipage}
    \caption{\textbf{Qualitative comparison on Composite Degradation.}}
    \label{fig:composite_vis}
\end{figure}

\begin{table}[t]
    \centering
    \caption{Quantitative results for \textbf{Composite Degradations}. Results taken from AgenticIR~\cite{agenticir} }
    \label{tab:all-in-one-models}
    \resizebox{0.48\textwidth}{!}{
    \begin{tabular}{ccccccc}
        \toprule[1.3pt]
        \textbf{Method} & \textbf{PSNR} $\uparrow$ & \textbf{SSIM} $\uparrow$  & \textbf{LPIPS}\,$\downarrow$ & \textbf{MANIQA}$\uparrow$  & \textbf{CLIP-IQA}$\uparrow$  & \textbf{MUSIQ}$\uparrow$  \\
        \midrule
        \rowcolor{gray!10} \multicolumn{7}{c}{\textbf{Group A}} \\
        \hline
            AirNet     & 19.13          & 0.6019          & 0.4283          & 0.2581          & 0.3930          & 42.46 \\
            PromptIR   & 20.06          & 0.6088          & 0.4127          & 0.2633          & 0.4013          & 42.62 \\
            MiOIR      & \underline{20.84}  & 0.6558      & 0.3715          & 0.2451          & 0.3933          & 47.82 \\
            DA-CLIP    & 19.58          & 0.6032          & 0.4266          & 0.2418          & 0.4139          & 42.51 \\
            InstructIR & 18.03          & 0.5751          & 0.4429          & 0.2660          & 0.3528          & 45.77 \\
            AdaIR      & 19.06          & 0.6061          & 0.4170          & 0.2718          & 0.4117          & 44.66 \\
            DARCOT     & 18.82          & 0.5960          & 0.4303          & 0.2641          & 0.4058          & 43.42 \\
            AutoDIR    & 19.64          & 0.6286          & 0.3967          & 0.2500          & 0.3767          & 47.01 \\
            AgenticIR  & \textbf{21.04} & \underline{0.6818}  & \underline{0.3148}  & \underline{0.3071}  & \underline{0.4474}  & \underline{56.88} \\
            \cellcolor{cyan!12}\method{} & \cellcolor{cyan!12}20.46 & \cellcolor{cyan!12}\textbf{0.7144} & \cellcolor{cyan!12}\textbf{0.1299} & \cellcolor{cyan!12}\textbf{0.4659} & \cellcolor{cyan!12}\textbf{0.6566} & \cellcolor{cyan!12}\textbf{57.19} \\
        \midrule
        \rowcolor{gray!10} \multicolumn{7}{c}{\textbf{Group B}} \\
        \hline
            AirNet     & 19.31          & 0.6567          & 0.3670          & 0.2882          & 0.4274          & 47.88 \\
            PromptIR   & 20.47          & 0.6704          & 0.3370          & 0.2893          & 0.4289          & 48.10 \\
            MiOIR      & \underline{20.56}  & 0.6905          & 0.3243          & 0.2638          & 0.4330          & 51.87 \\
            DA-CLIP    & 18.56          & 0.5946          & 0.4405          & 0.2435          & 0.4154          & 43.70 \\
            InstructIR & 18.34          & 0.6235          & 0.4072          & 0.3022          & 0.3790          & 50.94 \\
            AdaIR      & 19.24          & 0.6637          & 0.3467          & 0.3001          & 0.4380          & 49.16 \\
            DARCOT     & 19.08          & 0.6372          & 0.3910          & 0.2845          & 0.4321          & 46.37 \\
            AutoDIR    & 19.90          & 0.6643          & 0.3542          & 0.2534          & 0.3986          & 49.64 \\
            AgenticIR  & 20.55          & \underline{0.7009}  & \underline{0.3072}  & \underline{0.3204} & \underline{0.4648} & \textbf{57.57} \\
            \cellcolor{cyan!12}\method{} & \cellcolor{cyan!12}\textbf{21.04} & \cellcolor{cyan!12}\textbf{0.7326} & \cellcolor{cyan!12}\textbf{0.1269} & \cellcolor{cyan!12}\textbf{0.4582} & \cellcolor{cyan!12}\textbf{0.6483} & \cellcolor{cyan!12}\underline{56.91} \\
        \midrule
        \rowcolor{gray!10} \multicolumn{7}{c}{\textbf{Group C}} \\
        \hline
            AirNet     & 17.95          & 0.5145          & 0.5782          & 0.1854          & 0.3113          & 30.12 \\
            PromptIR   & 18.51          & 0.5166          & 0.5756          & 0.1906          & 0.3104          & 29.71 \\
            MiOIR      & 15.63          & 0.4896          & 0.5376          & 0.1717          & 0.2891          & 37.95 \\
            DA-CLIP    & 18.53          & 0.5320          & 0.5335          & 0.1916          & 0.3476          & 33.87 \\
            InstructIR & 17.09          & 0.5135          & 0.5582          & 0.1732          & 0.2537          & 33.69 \\
            AdaIR  & 17.88 & 0.5209 & 0.5802 & 0.2036 & 0.3276 & 30.83 \\
            DARCOT & 17.45 & 0.4993 & 0.6097 & 0.1937 & 0.3172 & 29.50 \\
            AutoDIR    & 18.61          & 0.5443          & 0.5019          & 0.2045          & 0.2939          & 37.86 \\
            AgenticIR  & \underline{18.82}  & \underline{0.5474}  & \underline{0.4493}  & \underline{0.2698}  & \underline{0.3948}  & \underline{48.68} \\
            \cellcolor{cyan!12}\method{} & \cellcolor{cyan!12}\textbf{19.33} & \cellcolor{cyan!12}\textbf{0.6579 }& \cellcolor{cyan!12}\textbf{0.1489} & \cellcolor{cyan!12}\textbf{0.4653} & \cellcolor{cyan!12}\textbf{0.6554} & \cellcolor{cyan!12}\textbf{56.56} \\
        \bottomrule[1.3pt]
    \end{tabular}}
\end{table}

\subsection{Comparison to state-of-the-art}
\label{sec:compare}

\begin{figure*}[t!]
  \centering
  \setlength{\tabcolsep}{1pt}
    \footnotesize
    \begin{tabular}{ccccccc}
        \includegraphics[width=.14\textwidth, height=2.0cm]{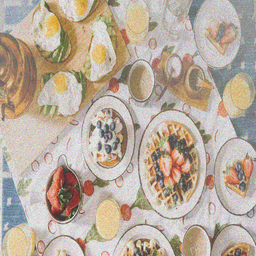} &
        \includegraphics[width=.14\textwidth, height=2.0cm]{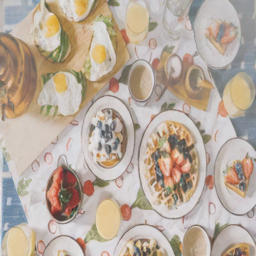} &
        \includegraphics[width=.14\textwidth, height=2.0cm]{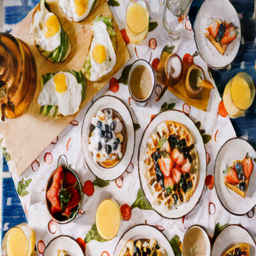} &
        \includegraphics[width=.14\textwidth, height=2.0cm]{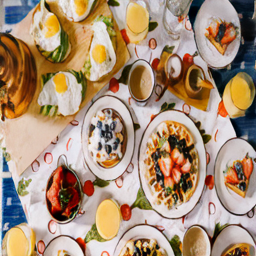} &
        \includegraphics[width=.14\textwidth, height=2.0cm]{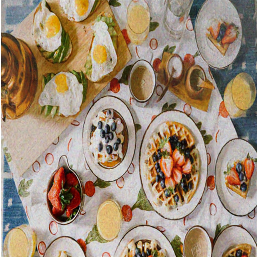} &
        \includegraphics[width=.14\textwidth, height=2.0cm]{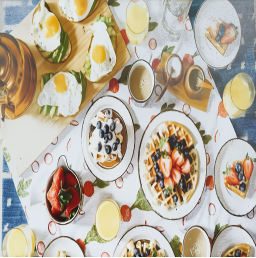} &
        \includegraphics[width=.14\textwidth, height=2.0cm]{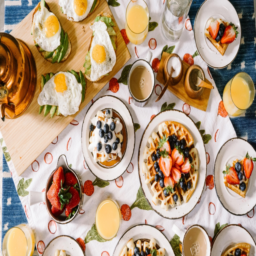} \\
        Step 1: \textcolor{blue}{noise} $\rightarrow$ & Step 2: \textcolor{blue}{haze} $\rightarrow$ & Step 3: \textcolor{blue}{contrast} $\rightarrow$ & \method{} & AutoDIR & AgenticIR & GT  \\
        \includegraphics[width=.14\textwidth, height=2.0cm]{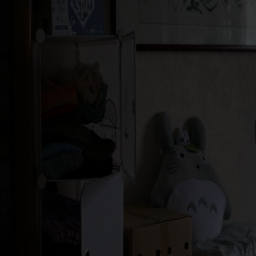} &
        \includegraphics[width=.14\textwidth, height=2.0cm]{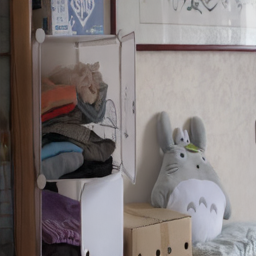} &
        \includegraphics[width=.14\textwidth, height=2.0cm]{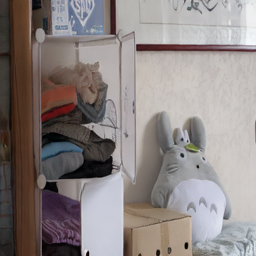} &
        \includegraphics[width=.14\textwidth, height=2.0cm]{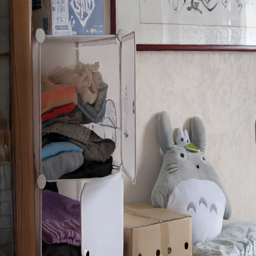} &
        \includegraphics[width=.14\textwidth, height=2.0cm]{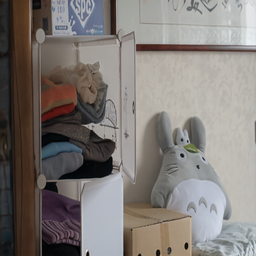} &
        \includegraphics[width=.14\textwidth, height=2.0cm]{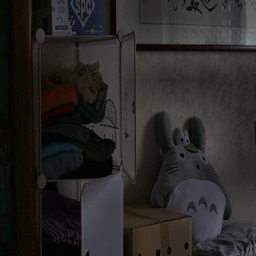} &
        \includegraphics[width=.14\textwidth, height=2.0cm]{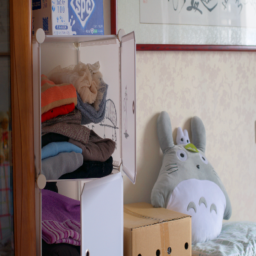} \\
        Step 1: \textcolor{blue}{low light} $\rightarrow$ & Step 2: \textcolor{blue}{blur} $\rightarrow$ & Step 3: \textcolor{blue}{none} $\rightarrow$ & \method{} & AutoDIR & AgenticIR & GT  \\
        \includegraphics[width=.14\textwidth, height=2.0cm]{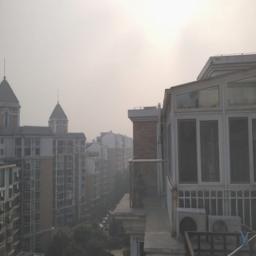} &
        \includegraphics[width=.14\textwidth, height=2.0cm]{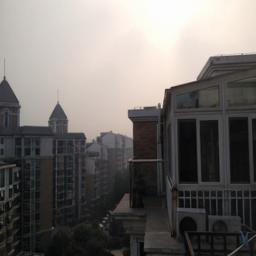} &
        \includegraphics[width=.14\textwidth, height=2.0cm]{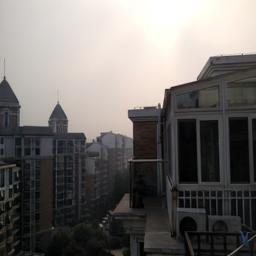} &
        \includegraphics[width=.14\textwidth, height=2.0cm]{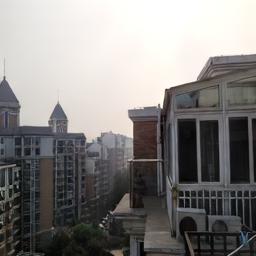} &
        \includegraphics[width=.14\textwidth, height=2.0cm]{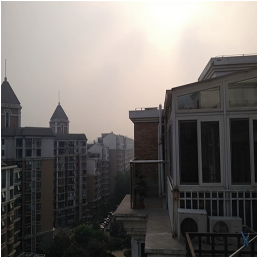} &
        \includegraphics[width=.14\textwidth, height=2.0cm]{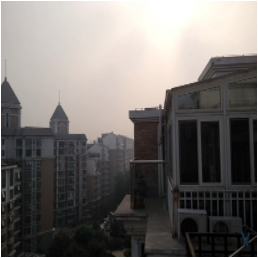} &
        \includegraphics[width=.14\textwidth, height=2.0cm]{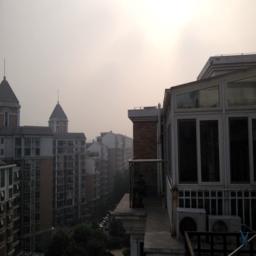} \\
        Step 1: \textcolor{blue}{haze} $\rightarrow$ & Step 2: \textcolor{blue}{low light} $\rightarrow$ & Step 3: \textcolor{blue}{low light} $\rightarrow$ & \method{} & AutoDIR & AgenticIR & GT  \\
        \includegraphics[width=.14\textwidth, height=2.0cm]{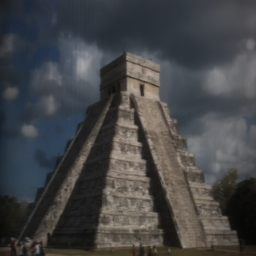} &
        \includegraphics[width=.14\textwidth, height=2.0cm]{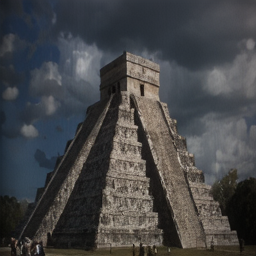} &
        \includegraphics[width=.14\textwidth, height=2.0cm]{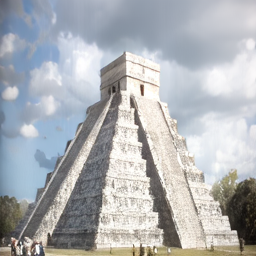} &
        \includegraphics[width=.14\textwidth, height=2.0cm]{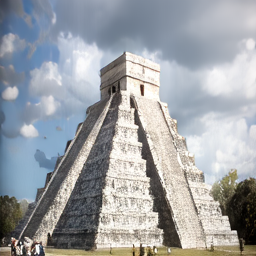} &
        \includegraphics[width=.14\textwidth, height=2.0cm]{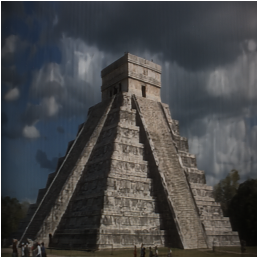} &
        \includegraphics[width=.14\textwidth, height=2.0cm]{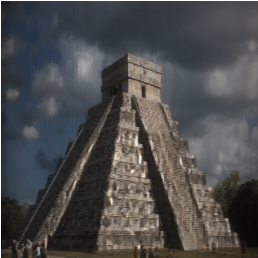} &
        \includegraphics[width=.14\textwidth, height=2.0cm]{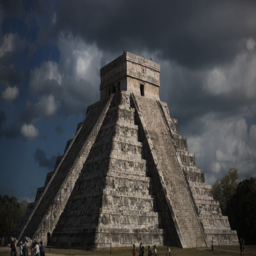} \\
        Step 1: \textcolor{blue}{blur} $\rightarrow$ & Step 2: \textcolor{blue}{low light} $\rightarrow$ & Step 3: \textcolor{blue}{saturation} $\rightarrow$ & \method{} & AutoDIR & AgenticIR  & GT  \\
    \end{tabular}
    \caption{\textbf{Qualitative analysis:} Comparison with other multi-round methods (AutoDIR~\cite{autodir}, AgenticIR~\cite{agenticir}). Intermediate steps for our method are shown for illustrative purposes. \textbf{First row:} Composite degradations (haze + noise). \method{} identifies and corrects them, also enhances contrast, making the result visually clearer. \textbf{Second row:} Single degradation (low light). AgenticIR fails due to inadequate tools for very dark scenes. \textbf{Third row:} Single degradation (haze). \method{} successfully dehazes, but also improves low light, lowering the fidelity to ground truth. \textbf{Fourth row:} Unknown degradations. \method{} demonstrates stronger generalization (also improving over ground truth).}
    \label{fig:iterative}
\end{figure*}

\noindent \textbf{Composite Degradations.}
We first evaluate our method in the composite degradation setup using the same test set introduced by AgenticIR~\cite{agenticir} and compare to main methods reporting results on this data. Table~\ref{tab:all-in-one-models}, speaks decisively in favor of our approach, which outperforms existing methods in almost all settings with \emph{significant} margins, where we are on average about $\times2$ better than state-of-the-art across perceptual metrics. Importantly, the superiority of our method is especially striking on the most challenging setting, which includes 3 degradations (i.e. Group C). Similarly, \cref{fig:composite_vis} shows that our method can generate richer details, making the images more realistic and better aligned with human visual perception. Also, the first row in Fig.~\ref{fig:iterative}, rolls out our iterative process on an example from this dataset. Clearly, our iterative process successfully identifies and fixes degradations to finally arrive at the cleanest image compared to strong baselines.\\

\begin{table}[t]
\centering
\caption{Results for \textbf{Unknown Degradations} on the UDC dataset.}
\label{tab:udc_zero_shot}
\resizebox{0.45\textwidth}{!}{%
\begin{tabular}{lcccc}
\toprule[1.3pt]
\textbf{Method} & \textbf{MUSIQ} $\uparrow$ & \textbf{CLIP-IQA} $\uparrow$ & \textbf{NIQE} $\downarrow$ & \textbf{NIMA} $\uparrow$ \\
\midrule
%
AirNet   & 41.20 & 0.265 & 7.681 & 4.048 \\
PromptIR & 45.40 & 0.303 & 8.294 & 4.344 \\
AdaIR    & 21.14 & 0.225 & 7.688 & 4.023 \\
DARCOT   & 20.38 & 0.211 & 8.224 & 4.039 \\
LD       & 43.41 & 0.288 & 7.983 & 4.371 \\
AutoDIR  & 49.27 & 0.311 & \underline{6.392} & 4.427 \\
AgenticIR & \underline{52.76} & \underline{0.358} & 6.670 & \underline{4.456} \\
\cellcolor{cyan!12}\method{} & \cellcolor{cyan!12}\textbf{55.97} & \cellcolor{cyan!12}\textbf{0.602}& \cellcolor{cyan!12}\textbf{6.162} & \cellcolor{cyan!12}\textbf{4.579} \\
\bottomrule[1.3pt]
\end{tabular}%
}
\vspace{-10pt}
\end{table}

\noindent \textbf{Unknown Degradations.} 
We follow previous work~\cite{autodir} and evaluate our method in the unknown degradation setup on the UDC~\cite{zhou2021image} dataset (e.g., TOLED) and use the perceptual metrics for this setup. As the results summarized in Table~\ref{tab:udc_zero_shot}, show that our method outperforms all baselines. Importantly, margins are especially sizable on the MUSIQ and CLIP-IQA metrics, which have been previously reported to align better with human judgment \cite{autodir}. The last row in Fig.~\ref{fig:iterative}, uses an image from this dataset. It provides an example of how our method can successfully handle unknown degradations, by iteratively removing degradations, yielding a final output that is even cleaner than the provided ground truth.\\

\begin{table}[t]
\centering
\caption{Quantitative comparison on \textbf{Single Degradation} tasks. }
\label{tab:single_distortion}
\resizebox{\columnwidth}{!}{%
\begin{tabular}{lcccccc}
\toprule[1.3pt]
\textbf{Method} & \textbf{PSNR} $\uparrow$ & \textbf{SSIM} $\uparrow$ & \textbf{LPIPS} $\downarrow$ & \textbf{CLIP-IQA} $\uparrow$ & \textbf{MUSIQ} $\uparrow$ & \textbf{MANIQA} $\uparrow$ \\
\midrule
AirNet   & 25.65 & 0.8230 & 0.1820 & \underline{0.4516} & 48.69 & 0.3163 \\
PromptIR & 27.04 & 0.8445 & 0.1515 & 0.4196 & 48.77 & 0.3169 \\
AdaIR    & 26.48 & 0.7825 & 0.2604 & 0.4368 & 53.51 & 0.3456 \\
DARCOT   & 26.08 & 0.7772 & 0.2579 & 0.4409 & 54.34 & \underline{0.3461} \\
LD       & 22.17 & 0.6922 & 0.1935 & 0.4253 & 54.94 & 0.3438 \\
AutoDIR  & \textbf{27.81} & \textbf{0.8703} & \underline{0.1283} & 0.4119 & 54.37 & 0.3053 \\
AgenticIR  & 24.41 & 0.7794 & 0.1392 & 0.3763 & \underline{55.95} & 0.2978 \\
\cellcolor{cyan!12}\method{} & \cellcolor{cyan!12}\underline{25.88} & \cellcolor{cyan!12}\underline{0.8378}& \cellcolor{cyan!12}\textbf{0.0699} & \cellcolor{cyan!12}\textbf{0.5566} & \cellcolor{cyan!12}\textbf{56.00} & \cellcolor{cyan!12}\textbf{0.4125} \\
\bottomrule[1.3pt]
\end{tabular}%
}
\vspace{-15pt}
\end{table}

\begin{figure*}[t!]
  \centering
  \setlength{\tabcolsep}{1pt}
    \footnotesize
    \begin{tabular}{cccccccc}
        \includegraphics[width=.12\textwidth, height=1.8cm]{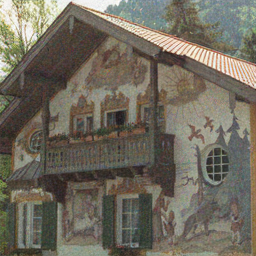} &
        \includegraphics[width=.12\textwidth, height=1.8cm]{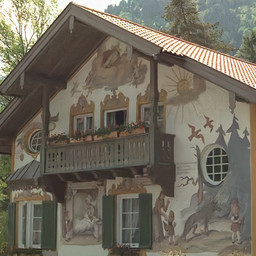} &
        \includegraphics[width=.12\textwidth, height=1.8cm]{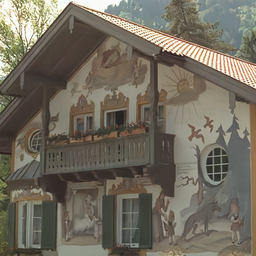} &
        \includegraphics[width=.12\textwidth, height=1.8cm]{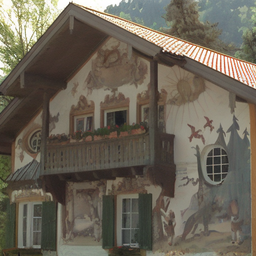} &
        \includegraphics[width=.12\textwidth, height=1.8cm]{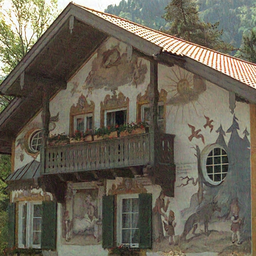} &
        \includegraphics[width=.12\textwidth, height=1.8cm]{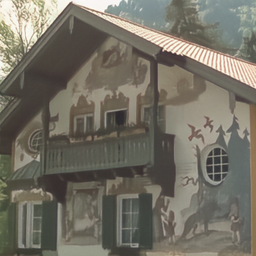} &
        \includegraphics[width=.12\textwidth, height=1.8cm]{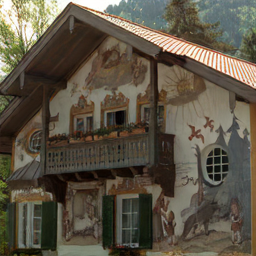} &
        \includegraphics[width=.12\textwidth, height=1.8cm]{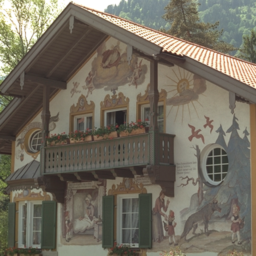} \\
        \includegraphics[width=.12\textwidth, height=1.8cm]{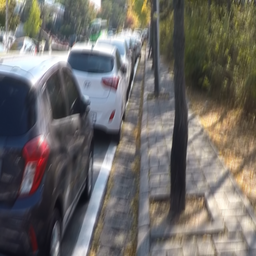} &
        \includegraphics[width=.12\textwidth, height=1.8cm]{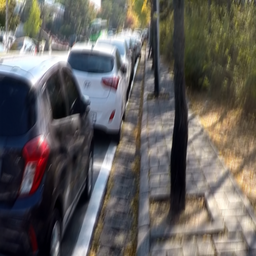} &
        \includegraphics[width=.12\textwidth, height=1.8cm]{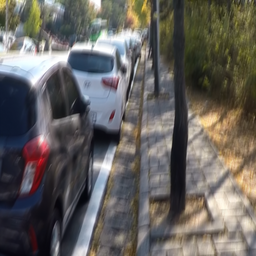} &
        \includegraphics[width=.12\textwidth, height=1.8cm]{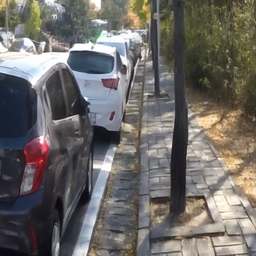} &
        \includegraphics[width=.12\textwidth, height=1.8cm]{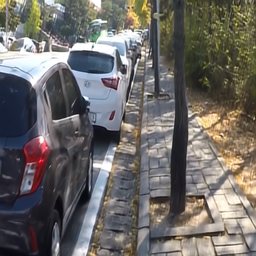} &
        \includegraphics[width=.12\textwidth, height=1.8cm]{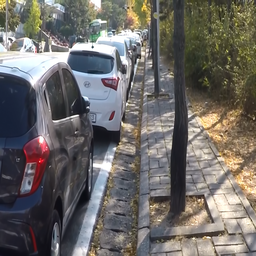} &
        \includegraphics[width=.12\textwidth, height=1.8cm]{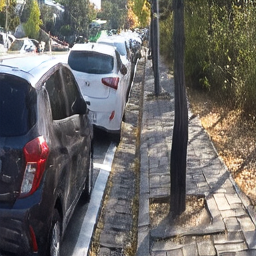} &
        \includegraphics[width=.12\textwidth, height=1.8cm]{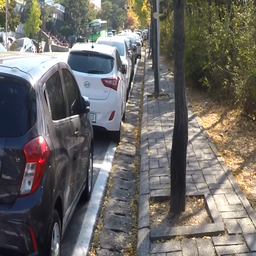} \\
        \includegraphics[width=.12\textwidth, height=1.8cm]{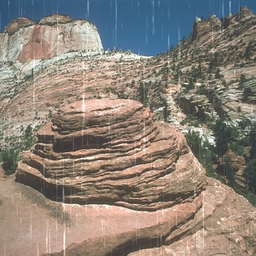} &
        \includegraphics[width=.12\textwidth, height=1.8cm]{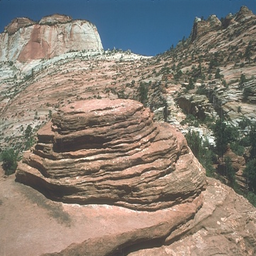} &
        \includegraphics[width=.12\textwidth, height=1.8cm]{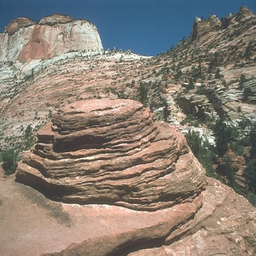} &
        \includegraphics[width=.12\textwidth, height=1.8cm]{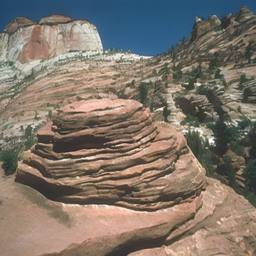} &
        \includegraphics[width=.12\textwidth, height=1.8cm]{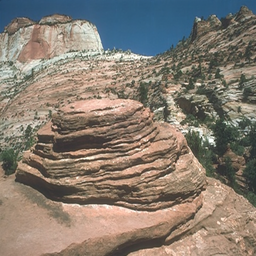} &
        \includegraphics[width=.12\textwidth, height=1.8cm]{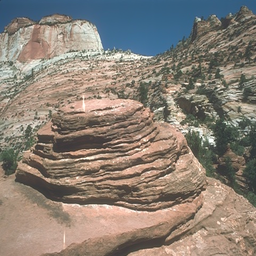} &
        \includegraphics[width=.12\textwidth, height=1.8cm]{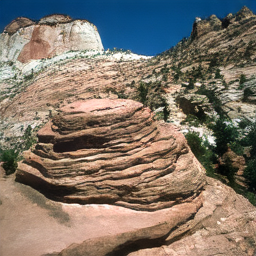} &
        \includegraphics[width=.12\textwidth, height=1.8cm]{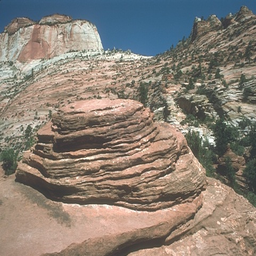} \\
        \includegraphics[width=.12\textwidth, height=1.8cm]{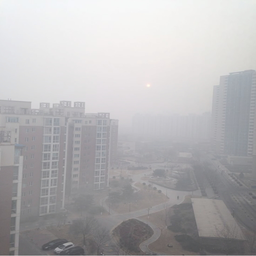} &
        \includegraphics[width=.12\textwidth, height=1.8cm]{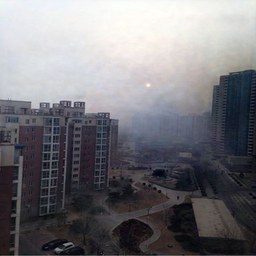} &
        \includegraphics[width=.12\textwidth, height=1.8cm]{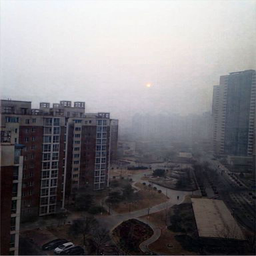} &
        \includegraphics[width=.12\textwidth, height=1.8cm]{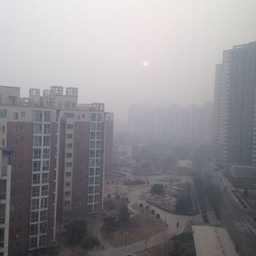} &
        \includegraphics[width=.12\textwidth, height=1.8cm]{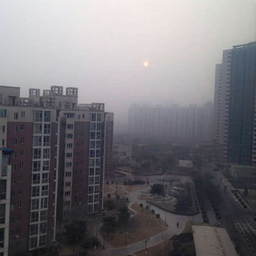} &
        \includegraphics[width=.12\textwidth, height=1.8cm]{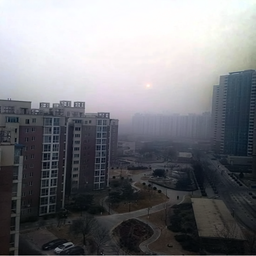} &
        \includegraphics[width=.12\textwidth, height=1.8cm]{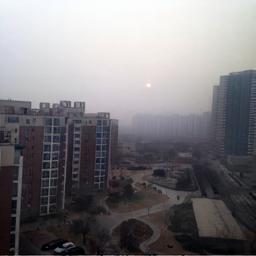} &
        \includegraphics[width=.12\textwidth, height=1.8cm]{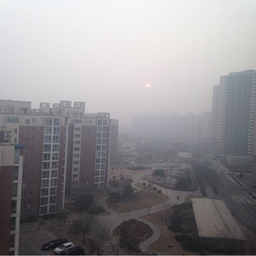} \\
        Input & AirNet & PromptIR & LD & AutoDIR & AgenticIR & \method{} & GT  \\
    \end{tabular}
    \caption{\textbf{Qualitative comparison on single degradation:} Row1: Kodak24 (noise), Row2: GoPro (blur), Row3: Rain100L (rain), Row4: SOTS (haze), In the single-degradation setting, our method produces clearer images that better align with human visual perception. This results lower fidelity scores but higher perceptual scores.}
    \label{fig:single_distortion}
\end{figure*}

\noindent \textbf{Single Degradation.} 
Finally, following AutoDIR~\cite{autodir}, we evaluate our performance on single degradation tasks. In particular, we evaluate on SIDD~\cite{abdelhamed2018high}, Kodak24~\cite{franzen1999kodak}, DIV2K~\cite{agustsson2017ntire}, GoPro~\cite{nah2017deep}, LOL~\cite{wei2018deep}, SOTS-Outdoor~\cite{li2018benchmarking}, Rain100~\cite{yang2017deep}, and Raindrop~\cite{qian2018attentive} datasets and report average scores across all tasks in Table~\ref{tab:single_distortion}. We perform particularly well on perceptual metrics in this setup. However, we note that our method is slightly behind on fidelity metrics. We attribute this behaviour to two main reasons. First, our iterative approach is particularly well suited for challenging multi-degradation setup. More importantly, we note that the ground truth only fixes one issue in the degraded image, even while there remains clear degradations (e.g. See GT in Fig.\ref{fig:single_distortion}). Therefore, our method, which tries to remove all degradations, goes beyond the GT, affecting fidelity metrics. For example, the middle two rows in Fig.~\ref{fig:iterative} show examples taken from single degradation test sets. Our model, can identify more degradations than advertised in GT and yield more visually appealing images. We provide more examples of this effect in the supplementary material.

\subsection{Ablation Study}\label{sec:ablation}
We ablate the components of our method in the challenging setup of unknown and composite degradations, using both fidelity (PSNR, SSIM) and perceptual metrics (LPIPS).

\vspace{4pt}
\noindent \textbf{Integration of the IQA and UIR modules.}
In Tab.~\ref{tab:abla_2} we ablate all steps in our method to tightly integrate the assessment and restoration modules. First, we focus on the role of the IQA model and integration of both image and text into the latent and embeddings space, respectively, of the restoration model. We follow the same setup as AutoDIR~\cite{autodir} in the first two rows to directly measure the impact of using a more descriptive IQA model. We use our flow-based formulation in the last four rows and show the great benefit of integrating the IQA into the restoration model latent space. Notably, replacing text with corresponding embeddings brings significant gains. We attribute this to the embeddings containing richer information than text, which is obtained from the embeddings but going through a lossy decoding step. This is illustrated in Fig. \ref{fig:abla_text_space}.

\begin{figure}[t]
  \centering
  \setlength{\tabcolsep}{1pt}
  \footnotesize
    \begin{tabular}{cccc}
        \includegraphics[width=.235\columnwidth]{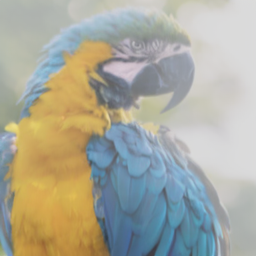} &
        \includegraphics[width=.235\columnwidth]{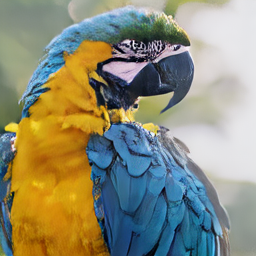} &
        \includegraphics[width=.235\columnwidth]{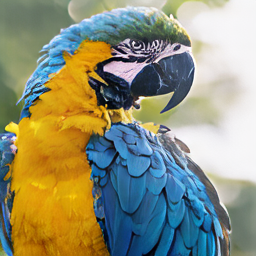} &        
        \includegraphics[width=.235\columnwidth]{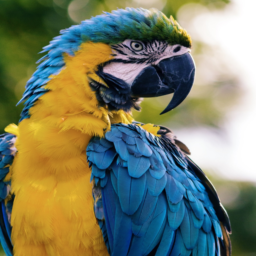} \\
        (a) Input & (b) SD3.5 & (c) Noise-free SD3.5 & (d) GT \\
    \end{tabular}
\caption{\textbf{Noise vs. noise-free flow matching formulation.} Notice how noise affects the eye of the parrot (zoom required). } 
    \label{fig:abla_noise}
\end{figure}

\begin{figure}[t]
  \centering
  \setlength{\tabcolsep}{1pt}
  \footnotesize
    \begin{tabular}{cccc}
        \includegraphics[width=.238\columnwidth]{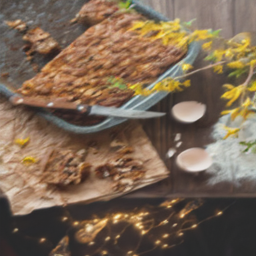} &
        \includegraphics[width=.238\columnwidth]{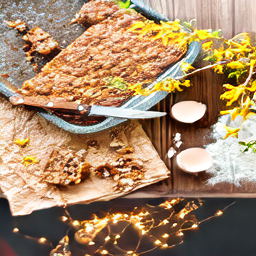} &
        \includegraphics[width=.238\columnwidth]{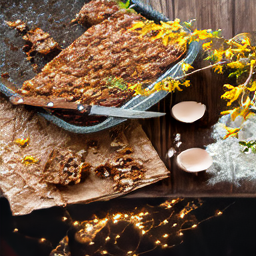} &        
        \includegraphics[width=.238\columnwidth]{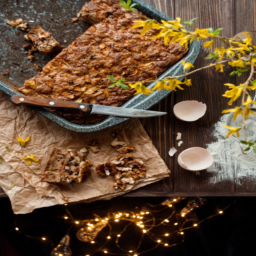} \\
        (a) Input & (b) Text & (c) Embeddings & (d) GT \\
    \end{tabular}
\caption{\textbf{Text vs. embeddings as conditioning inputs.} Embeddings provide more efficient and informative conditioning.
}
    \label{fig:abla_text_space}
\end{figure}

Switching attention to the role of the UIR backbone, we can see that while relying on the stronger SD3.5 brings a sizable performance boost for unknown degradations, its effect is minimal in the case of composite degradations, which shows that the task needs more than a higher capacity model. Importantly, removing the noise-based conditioning in favor of directly mapping from degraded to high quality image distributions does not adversely affect performance and yields better perceptual results as illustrated in Fig.~\ref{fig:abla_noise}. 

\begin{table*}[t]
    \centering
    \caption{\textbf{Ablation} of the integration process between the IQA and the UIR modules.}
    \label{tab:abla_2}
    \resizebox{\textwidth}{!}{
    \begin{tabular}{c|cccc|ccc|ccc}
    \toprule[1.3pt]
    \multirow{2}{*}{\textbf{Backbone}} & \multicolumn{4}{c|}{\textbf{Configuration}} & \multicolumn{3}{c|}{\textbf{Unknown Degradations (UDC)}} & \multicolumn{3}{c}{\textbf{Composite Degradations (Group-C)}} \\
     & IQA & Text Space & Image Space & noise-free cond. & \textbf{PSNR} $\uparrow$ & \textbf{SSIM} $\uparrow$ & \textbf{LPIPS} $\downarrow$ & \textbf{PSNR} $\uparrow$ & \textbf{SSIM} $\uparrow$ & \textbf{LPIPS} $\downarrow$ \\
    \midrule
    SD1.5  & CLIP     & Text  & Pixel & \xmark & 20.67 &	0.6217 &  0.2211 & 16.70 & 0.4848 & 0.2997 \\
    SD1.5  & DepictQA & Text  & Pixel & \xmark &22.34 & 0.6596 & 0.2016 & 17.67 & 0.5260 & 0.2267 \\
    \hline
    SD3.5 & DepictQA & Text  & Pixel & \xmark  &    24.41 & 0.7583 & 0.1652 & 17.88 & 0.5696 & 0.2306\\
    SD3.5  & DepictQA & Text  & Pixel & \cmark &  24.70 & 0.7766 & 0.1589 & 17.89 & 0.5776 & 0.2289 \\
    SD3.5  & DepictQA & Text  & Latent & \cmark & 24.90 & 0.7952 & 0.1131 & 18.72 & 0.6278 & 0.1576 \\
    \rowcolor{cyan!12} SD3.5  & DepictQA & Embedding & Latent & \cmark & 28.49 & 0.8494 & 0.1007 & 18.76 & 0.6254 & 0.1730 \\
    \bottomrule[1.3pt]
    \end{tabular}}
\end{table*}

\vspace{4pt}
\noindent \textbf{Role of the \method{} iterative process.} 
We now show the benefit the iterative process, enabled by our tight integration, both at training and inference time. In Table~\ref{tab:abla_3} we can see that using iterative conditioning at training time benefits the flow-based formulation, while it actually is detrimental to the diffusion-based model. We attribute this behaviour to the utilization of a noising process in the diffusion paradigm, which alters the latents consumed by the LQA, thereby making it less useful. In contrast, our noise-free formulation using flow-matching allows us to directly use the degraded latents during training without noise contamination, thereby taking better advantage of the LQA feedback at training time. Importantly, we note that this iterative conditioning benefits more the composite degradation setup as it is best aligned with our iterative setup.

\vspace{4pt}
\noindent\textbf{Effectiveness of the \method{} stopping criterion.}
Finally, we analyze the effectiveness of the proposed method for stopping the \method{} iterations. To this end we run the \method{} process for varying number of \emph{predefined} fixed iterations and compare results to directly using our stopping criterion. Given the time consuming nature of this experimental setup, we only report results on the UDC~\cite{zhou2021image} dataset in Tab.~\ref{tab:abla_4}. Using our stopping criterion, we average 2.4 \method{} rounds, which strikes a good balance between fidelity and perceptual metrics, demonstrating the efficacy of \method{}.

\vspace{4pt}
\noindent \textbf{Computational Efficiency.}\label{sec:efficiency}
In addition to qualitative and quantitative comparisons, we also wanted to highlight the efficiency of the proposed iterative process in terms of latency compared to other iterative methods. In particular, we compare to AgenticIR \cite{agenticir} and AutoDIR~\cite{autodir}. 
In Table~\ref{supp:tab:higher-threshold-time} we can see that our \method{} process is a lot more efficient both in terms of latency and average number of iterations required, which shows that our all-in-one model can readily compete, both in terms of performance and efficiency, with methods using a suite of tools and other all-in-one iterative models, even while they have lower parameter count. 

\begin{table}[t]
    \centering
    \caption{\textbf{Role of the \method{} iterative process.} Intermediate assessments are not accurate for SD1.5 due to the noised latents. Instead, our noise-free FM variant improves across all metrics and datasets.}
    \label{tab:abla_3}
    \resizebox{\columnwidth}{!}{
    \begin{tabular}{c|c|ccc|ccc}
    \toprule[1.3pt]
    \multirow{2}{*}{\textbf{Backbone}} & \multicolumn{1}{c|}{\textbf{Training approach}} & \multicolumn{3}{c|}{\textbf{Unknown (UDC)}} & \multicolumn{3}{c}{\textbf{Composite (Group-C)}} \\
     & Iterative & \textbf{PSNR} $\uparrow$ & \textbf{SSIM} $\uparrow$ & \textbf{LPIPS} $\downarrow$ & \textbf{PSNR} $\uparrow$ & \textbf{SSIM} $\uparrow$ & \textbf{LPIPS} $\downarrow$ \\
    \midrule
    SD1.5   &      -      & 25.57 & 0.747 & 0.132 & 18.37 & 0.555 & 0.175 \\
    SD1.5   & \checkmark & 23.05 & 0.697 & 0.154 & 18.17 & 0.553 & 0.177 \\
    \hline
    SD3.5   &       -     & 28.49 & 0.849 & 0.101 & 18.76 & 0.625 & 0.173 \\
    \rowcolor{cyan!12} SD3.5   & \checkmark & \textbf{28.60} & \textbf{0.856} & \textbf{0.083} & \textbf{19.16} & \textbf{0.649} & \textbf{0.149} \\
    \bottomrule[1.3pt]
    \end{tabular}}
\end{table}

\begin{table}[t]
    \centering
    \caption{\textbf{Stopping Criterion Analysis.} Results for unknown composite degradations on the UDC dataset. }
    \label{tab:abla_4}
    \resizebox{\columnwidth}{!}{
    \begin{tabular}{c|cccccc}
    \toprule[1.3pt]
    \# RAR rounds &  \textbf{PSNR} $\uparrow$ & \textbf{SSIM} $\uparrow$ & \textbf{LPIPS} $\downarrow$ & \textbf{CLIPIQA} $\uparrow$ & \textbf{MUSIQ} $\uparrow$ & \textbf{MINIQA} $\uparrow$ \\
    \midrule
    1 & 30.69 & 0.901 & 0.0583 & 0.591 & 54.87 & 0.417 \\ 
    2 & 29.55 & 0.873 & 0.073 & 0.601 & 56.29 & 0.428 \\ 
    3 & 29.12 & 0.866 & 0.077 & 0.608 & 56.80 & 0.438 \\ 
    4 & 28.60 & 0.856 & 0.083 & 0.614 & 57.19 & 0.446 \\ 
    \rowcolor{cyan!12} 2.4 & 29.69 & 0.874 & 0.071 & 0.603 & 55.97 & 0.423 \\ 
    \bottomrule[1.3pt]
    \end{tabular}}
\end{table}

\begin{table}[t]
    \centering
    \caption{\textbf{Analysis of Computational Efficiency.} Comparison with iterative approaches in terms of latency and performance.}
    \label{supp:tab:higher-threshold-time}
    \resizebox{\columnwidth}{!}{
    \begin{tabular}{clccccc}
    \toprule[1.3pt]
    Degradations & Method & Wall clock time & \# of Round & \textbf{PSNR} $\uparrow$ & \textbf{SSIM} $\uparrow$ & \textbf{LPIPS}\,$\downarrow$\\ 
    \midrule
    \multirow{2}{*}{Group A} & AutoDIR & 14.30 & 2.92 & 19.64 & 0.6286 & 0.3967 \\
                             & AgenticIR & 48.00 & 3.37  & \textbf{21.04} & 0.6818 & 0.3148 \\
                             & \cellcolor{cyan!12}\method{} & \cellcolor{cyan!12}\textbf{6.29} & \cellcolor{cyan!12}\textbf{2.82} & \cellcolor{cyan!12}20.46 & \cellcolor{cyan!12}\textbf{0.7144} & \cellcolor{cyan!12}\textbf{0.1299} \\
    \midrule
    \multirow{2}{*}{Group B} & AutoDIR & 10.59 & \textbf{1.98} & 19.90 & 0.6643 & 0.3542 \\
                             & AgenticIR & 54.00 & 3.63  & 20.55 & 0.7009 & 0.3072 \\
                             & \cellcolor{cyan!12}\method{} & \cellcolor{cyan!12}\textbf{6.71} & \cellcolor{cyan!12}2.84 & \cellcolor{cyan!12}\textbf{21.04} & \cellcolor{cyan!12}\textbf{0.7326} & \cellcolor{cyan!12}\textbf{0.1269}\\
    \midrule
    \multirow{2}{*}{Group C} & AutoDIR & 18.12 & 4.42 & 18.61 & 0.5443 & 0.5019 \\
                             & AgenticIR & 78.00 & 4.77 & 18.82 & 0.5474 & 0.4493 \\
                             & \cellcolor{cyan!12}\method{} & \cellcolor{cyan!12}\textbf{6.92} & \cellcolor{cyan!12}\textbf{2.94}  & \cellcolor{cyan!12}\textbf{19.33} & \cellcolor{cyan!12}\textbf{0.6579} & \cellcolor{cyan!12}\textbf{0.1489}\\
    \bottomrule[1.3pt]
    \end{tabular}}
\end{table}
\section{Conclusion}

We introduce a new methodology for image restoration that can seamlessly handle unknown composite degradations. Our method starts with aligning the tasks of free-form IQA and all-in-one restoration to form a unified model that can both assess and restore. We further improve upon this design by tightly integrating both components into a single end-to-end trainable model, where the inter-dependencies of both tasks can be fully exploited. Finally, we enable adaptive assessment, both at inference and training time, leading to our restore, assess, repeat model. Our method establishes a new state-of-the-art across all datasets considered, including single, composite and unknown degradations, all under one single test-time configuration. 
\clearpage

\appendix

\begin{center}
\textbf{\large Supplemental Materials}
\end{center}
\renewcommand{\theequation}{S.\arabic{equation}}
\renewcommand\thefigure{S.\arabic{figure}}
\renewcommand\thetable{S.\arabic{table}}
\setcounter{equation}{0}
\setcounter{figure}{0}
\setcounter{table}{0}
\section{Latent Quality Assessment}

\subsection{Finetuning and Usage}
\begin{figure}[t]
  \centering
  \setlength{\tabcolsep}{1pt}
    \footnotesize
    \includegraphics[width=\columnwidth]{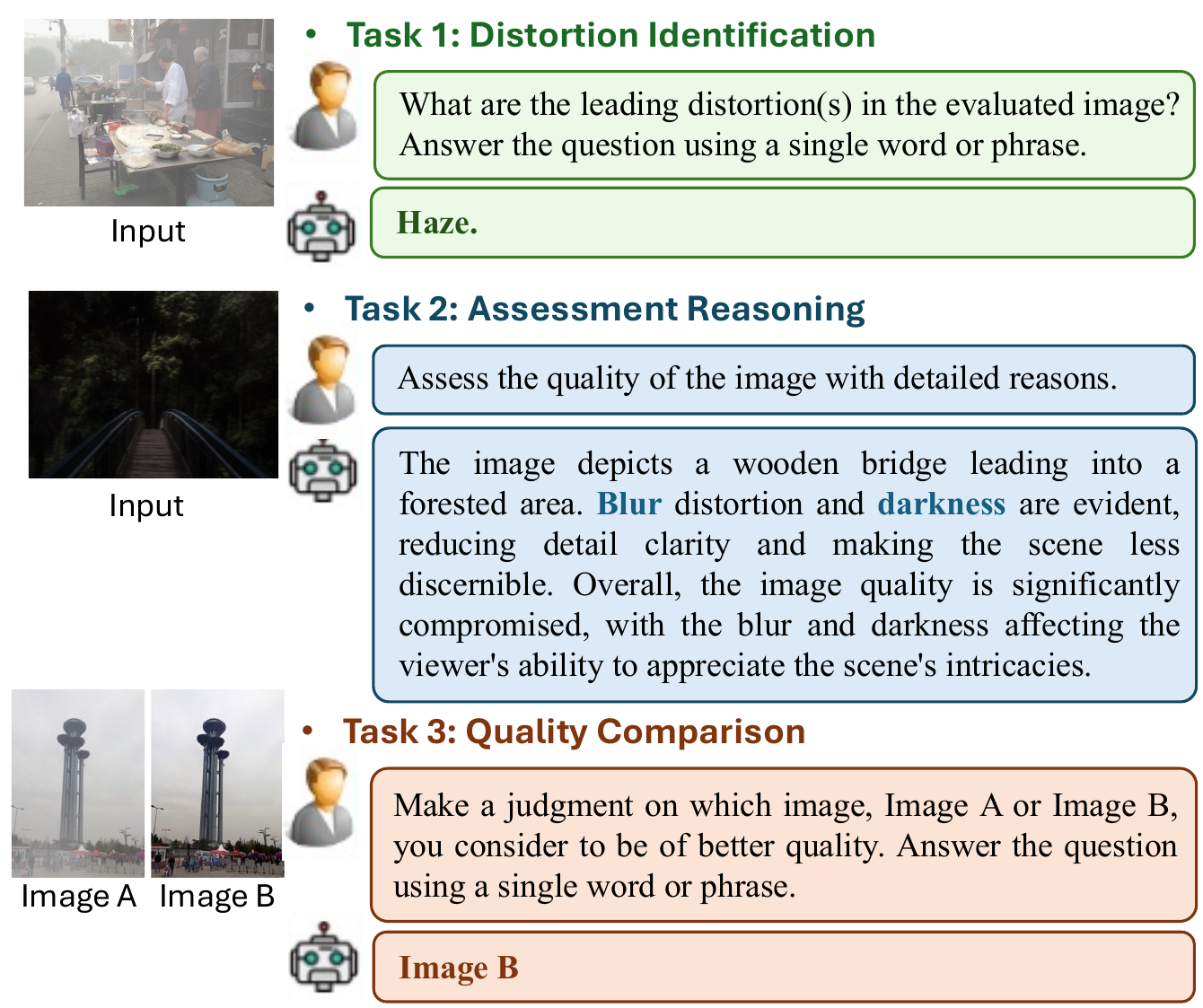} 
    \caption{\textbf{Illustrations of the three prompt-driven IQA tasks used in LQA \emph{training}}: distortion identification, assessment reasoning, and quality comparison. These tasks collectively preserve rich degradation-awareness within the unified VLM backbone.}
    \label{fig:LQA_tasks}
\end{figure}

To enable rich degradation-awareness of the LQA module, we follow the DepictQA~\cite{depictqa} pipeline and further finetune the original model on three complementary tasks: (1) distortion identification, (2) assessment reasoning, and (3) quality comparison, as illustrated in~\figref{fig:LQA_tasks}. To better adapt LQA to our restoration setting, we further extend the distortion-identification corpus using all training pairs from the restoration datasets. During inference, we primarily use distortion identification to generate conditioning signals, and quality comparison to determine the RAR stopping criterion, all using the same VLM backbone. 

\subsection{Architecture Details}
To unify the IQA and restoration modules, we introduced two components: latent space alignment and conditioning alignment, enabling the assessment model to operate directly in the shared latent domain. We describe each component below.

\begin{figure}[t]
  \centering
  \setlength{\tabcolsep}{1pt}
    \footnotesize
    \includegraphics[width=\columnwidth]{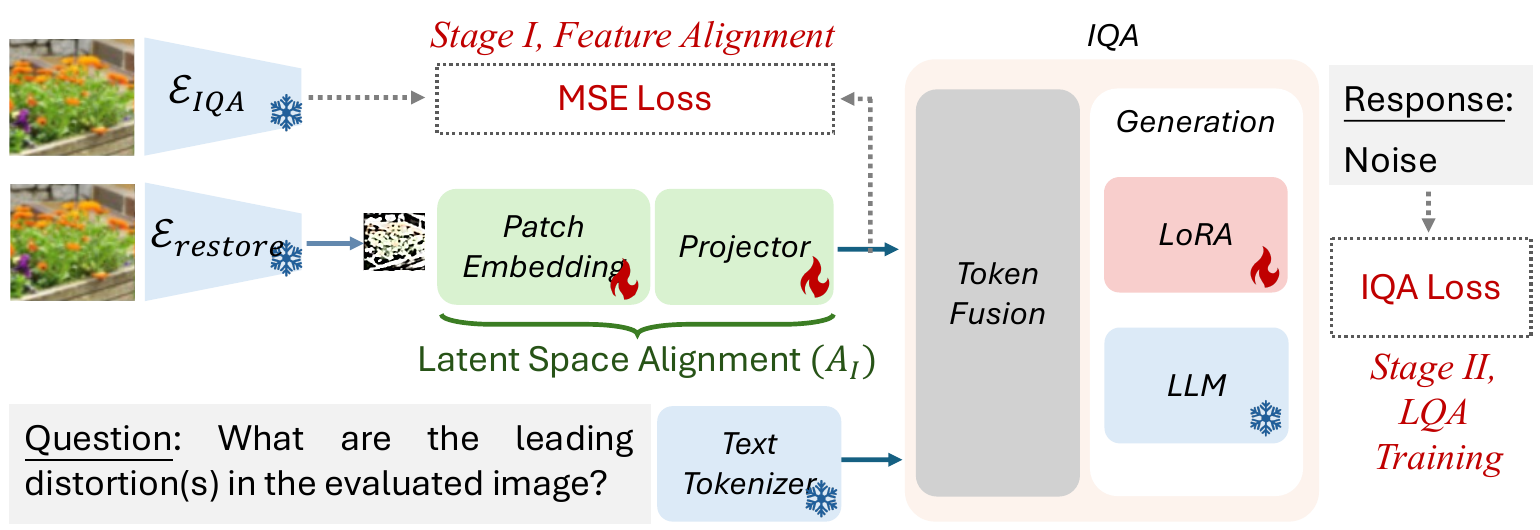} 
    \caption{\textbf{Overview of the latent space alignment module}, which allows us to use a single encoder for both IQA and restoration modules.}
    \label{fig:LQA_feat_align}
\end{figure}

\noindent\textbf{Latent Space Alignment}
Latent space alignment enables the IQA module to operate directly within the shared latent space of the restoration model. Instead of using the original IQA encoder $\mathcal{E}_{IQA}$, we replace it with the restoration encoder $\mathcal{E}_{restore}$ and introduce an image adapter $A_I$ to bridge their feature spaces. Training proceeds in two stages: first, we align the adapter outputs to the IQA encoder representations using simple MSE loss; then, we perform end-to-end finetuning, following the original IQA losses. An overview is shown in~\figref{fig:LQA_feat_align}.

\noindent\textbf{Conditioning Alignment}
Stable Diffusion~3.5~\cite{esser2024scaling} relies on three separate text encoders (CLIP-L~\cite{radford2021learning}, CLIP-G~\cite{radford2021learning}, and T5-XXL~\cite{raffel2020exploring}) to produce conditioning embeddings, making text-based conditioning computationally expensive and inefficient for iterative restoration. Moreover, the IQA module naturally outputs free-form text, which would require decoding and subsequent re-encoding through these large text encoders in our target setup, a process that introduces unnecessary information loss.

\begin{figure}[t]
  \centering
  \setlength{\tabcolsep}{1pt}
    \footnotesize
    \includegraphics[width=\columnwidth]{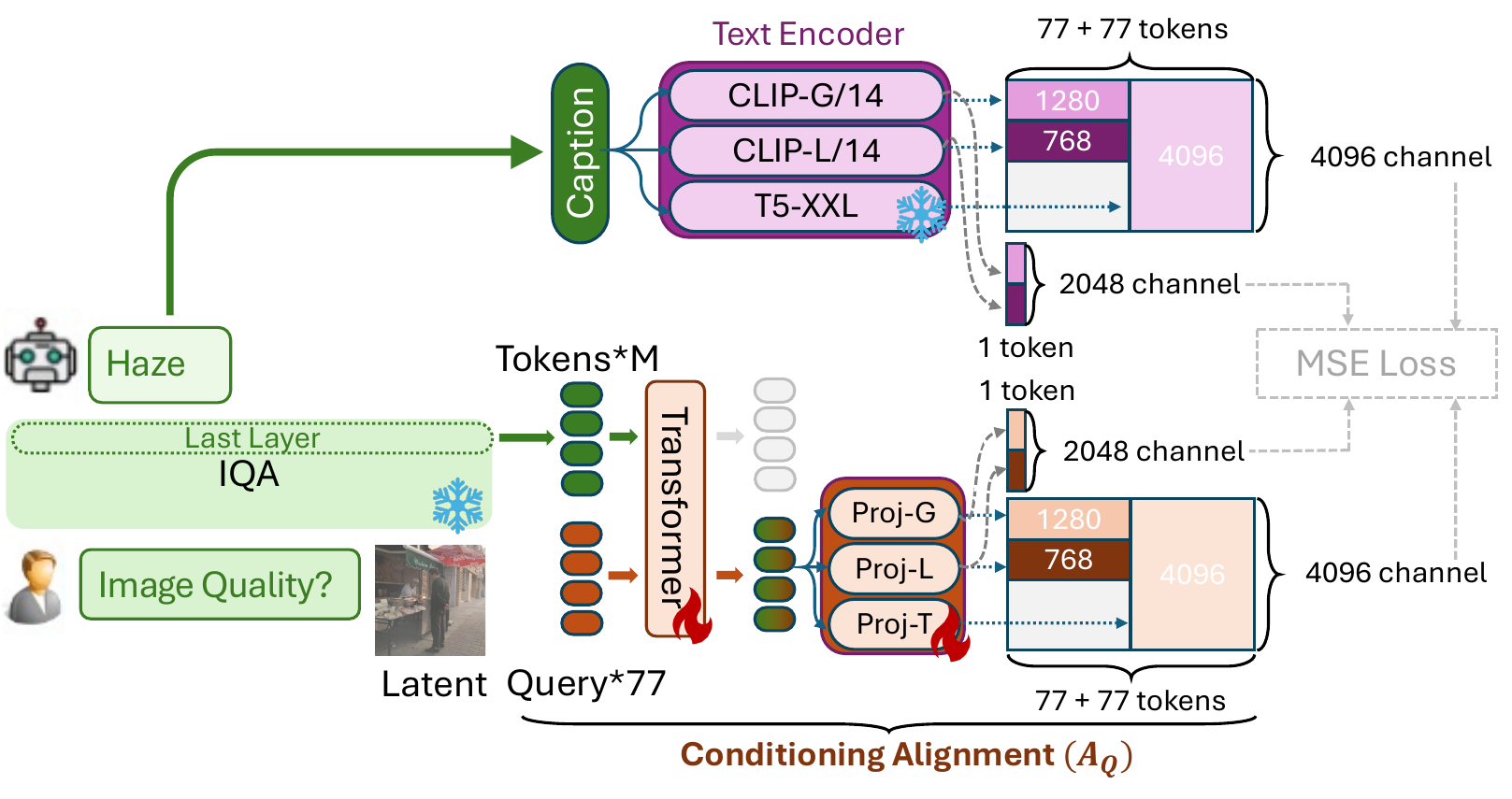} 
    \caption{\textbf{Overview of the conditioning alignment module}, which maps LQA outputs into the UIR conditioning space.}
    \label{fig:LQA_cond_align}
\end{figure}

Instead, we project the IQA output latent 
$\tilde{Q}_{\text{deg}}$ into the UIR conditioning space through an adapter $A_Q$, which is trained to match the embeddings produced by the restoration model's text-conditioning branch. The alignment is optimized in two stages: first finetuning $A_Q$, and then jointly training it with the UIR module, as shown in~\figref{fig:LQA_cond_align}.

This design eliminates the need for IQA decoding and text re-encoding, thereby avoiding both information loss and computational inefficiency, while enabling fully end-to-end conditioning.

\begin{table}[t]
    \centering
    \caption{\textbf{Distortion identification on composite degradations.} Evaluated on KADIS700K~\cite{lin2020deepfl} dataset.}
    \label{tab:lqa_iden1}
    \resizebox{\columnwidth}{!}{
    \begin{tabular}{c|cccc}
    \toprule[1.3pt]
    Method & Precision $\uparrow$ & Recall $\uparrow$ & F1-score $\uparrow$ \\
    \midrule
    DepictQA (Pixel-based IQA) & 0.7713 & 0.7564 & 0.7612 \\ 
    \rowcolor{cyan!12} LQA & \textbf{0.8750} & \textbf{0.8748} & \textbf{0.8749} \\ 
    \bottomrule[1.3pt]
    \end{tabular}}
\end{table}

\begin{table}[t]
    \centering
    \caption{\textbf{Distortion identification on single degradations.} F1-scores on each restoration dataset.}
    \label{tab:lqa_iden2}
    \resizebox{\columnwidth}{!}{
    \begin{tabular}{c|cccccccc}
    \toprule[1.3pt]
    Method & Haze & Blur & Rain & LOL & Raindrop & Noise & Low-Res & Average \\
    \midrule
    CLIP & 0.674 & 0.721 & 0.889 & 0.882 & 0.757 & 0.805 & 0.723 & 0.778 \\ 
    SA-CLIP & 1.000 & 0.944 & 0.995 & 1.000 & 1.000 & 1.000 & 0.998 & 0.991 \\ 
    \rowcolor{cyan!12} LQA & \textbf{1.000} & \textbf{1.000} & \textbf{1.000} & \textbf{1.000} & \textbf{1.000} & \textbf{1.000} & \textbf{0.975} & \textbf{0.996} \\ 
    \bottomrule[1.3pt]
    \end{tabular}}
\end{table}

\begin{table}[t]
    \centering
    \caption{\textbf{Sensitivity Analysis of User Prompts.}}
    \label{tab:prompt_test}
    \resizebox{\columnwidth}{!}{
    \begin{tabular}{lccccccc}
        \toprule
        Degradation & Prompts &  PSNR $\uparrow$ & SSIM $\uparrow$ & LPIPS $\downarrow$ & CLIPIQA $\uparrow$ & MUSIQ $\uparrow$ & MINIQA $\uparrow$ \\
        \hline
        \multirow{2}{*}{Group C}
        & Fixed     & 19.33 & 0.6579 & 0.1489 & 0.6554 & 56.56 & 0.4653 \\  
        & Random    & 19.31 & 0.6572 & 0.1493 & 0.6566 & 56.72 & 0.4675 \\ 
        \bottomrule
    \end{tabular}
    }
\end{table}

\subsection{Evaluation of LQA}
To validate the accuracy of our LQA module, we evaluate distortion identification performance on both composite and single-degradation settings. On the publicly available KADIS700K~\cite{lin2020deepfl} composite degradation benchmark, LQA significantly outperforms the original DepictQA model. We further compare against CLIP~\cite{esser2024scaling} and SA-CLIP (as adopted in AutoDIR~\cite{autodir}) on the restoration datasets, where LQA achieves near-perfect F1-scores across all degradation types. These results demonstrate that LQA preserves strong degradation-awareness while operating entirely in the shared latent space.

\subsection{Prompt Definition}
We follow the DepictQA~\cite{depictqa} procedure and use a fixed prompt template in our LQA. The exact prompt formulation is provided in~\figref{fig:LQA_tasks}. To evaluate sensitivity to prompt wording, we additionally construct a predefined pool of 24 semantically equivalent prompts and randomly sample one at test time for each decision step. Results on composite distortion-Group C are summarized in~\tabref{tab:prompt_test}. As shown, the performance remains highly consistent between the fixed-prompt and random-prompt settings across all full-reference and no-reference metrics. These results indicate that the stopping decision is not sensitive to minor variations in prompt wording, suggesting that the LQA-based termination mechanism provides a stable decision signal.
\section{More Experimental Results}

\begin{figure}[t]
  \centering
  \setlength{\tabcolsep}{1pt}
    \footnotesize
    \includegraphics[width=\columnwidth]{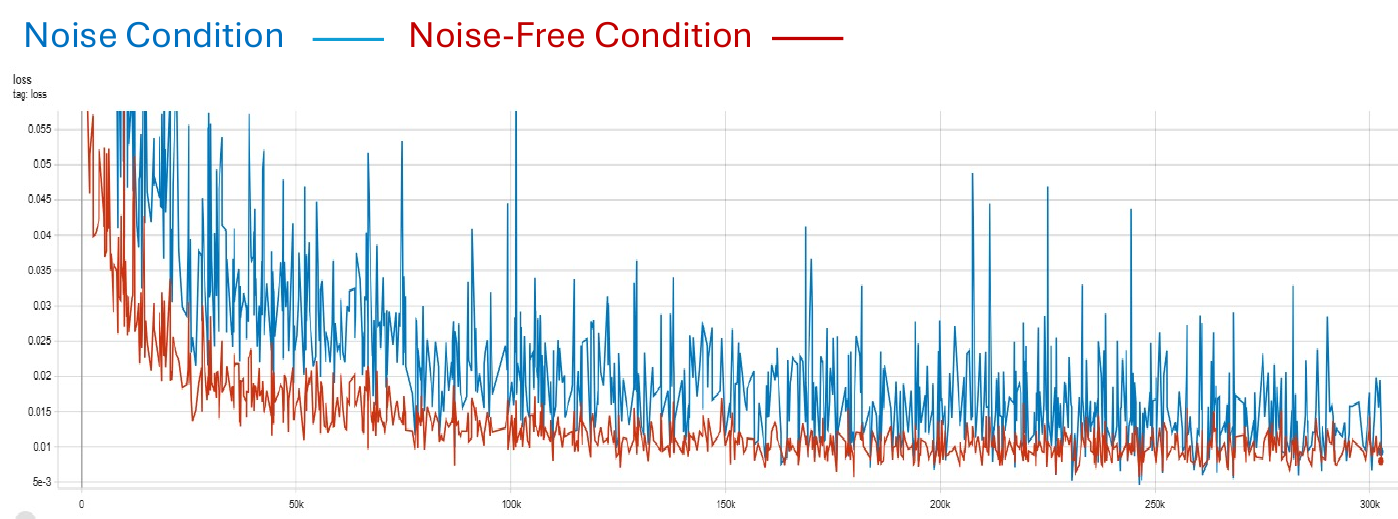} 
    \caption{Training curves comparing noise-free flow matching with the standard noise-conditioned variant, showing significantly faster and stable convergence.}
    \label{fig:flow_matching_curve}
\end{figure}

\begin{table*}[t]
    \centering
    \caption{\textbf{Ablation} of the integration process between the IQA and the UIR modules.}
    \label{tab:abla_supp}
    \resizebox{\textwidth}{!}{
    \begin{tabular}{c|cccc|ccc|ccc}
    \toprule[1.3pt]
    \multirow{2}{*}{\textbf{Backbone}} & \multicolumn{4}{c|}{\textbf{Configuration}} & \multicolumn{3}{c|}{\textbf{Unknwon (UDC)}} & \multicolumn{3}{c}{\textbf{Composite (Group-C)}} \\
     & IQA & Text Space & Image Space & Iterative Training & \textbf{PSNR} $\uparrow$ & \textbf{SSIM} $\uparrow$ & \textbf{LPIPS} $\downarrow$ & \textbf{PSNR} $\uparrow$ & \textbf{SSIM} $\uparrow$ & \textbf{LPIPS} $\downarrow$ \\
    \midrule
    AutoDIR& CLIP     & Text       & Pixel  & \xmark & 22.47 & 0.7267 & 0.2199 & 18.61 & 0.5443 & 0.5019 \\
    \hdashline
    SD1.5  & CLIP     & Text       & Pixel  & \xmark & 20.67 & 0.6217 & 0.2211 & 16.70 & 0.4848 & 0.2997 \\
    \hdashline
    SD1.5  & DepictQA & Text       & Pixel  & \xmark & 22.34 & 0.6596 & 0.2016 & 17.67 & 0.5260 & 0.2267 \\
    SD1.5  & DepictQA & Text       & Latent & \xmark & 23.66 & 0.7055 & 0.1498 & 18.06 & 0.5475 & 0.1792 \\
    SD1.5  & DepictQA & Embeddings & Latent & \xmark & 25.57 & 0.7473 & 0.1319 & 18.37 & 0.5546 & 0.1751 \\
    SD1.5  & DepictQA & Embeddings & Latent & \cmark & 23.05 & 0.6970 & 0.1544 & 18.17 & 0.5530 & 0.1774 \\
    \hline
    SD3.5  & DepictQA & Text  & Pixel    &  \xmark & 24.70 & 0.7766 & 0.1589 & 17.89 & 0.5776 & 0.2289 \\
    SD3.5  & DepictQA & Text  & Latent   & \xmark  & 24.90 & 0.7952 & 0.1131 & 18.72 & 0.6278 & 0.1576 \\
    SD3.5  & DepictQA & Embedding & Latent &\xmark & 28.49 & 0.8494 & 0.1007 & 18.76 & 0.6254 & 0.1730 \\
    \rowcolor{cyan!12} SD3.5  & DepictQA & Embedding & Latent & \cmark & \textbf{28.60} & \textbf{0.8559} & \textbf{0.0826} & \textbf{19.16} & \textbf{0.6494} & \textbf{0.1495} \\
    \bottomrule[1.3pt]
    \end{tabular}}
    \vspace{-10pt}
\end{table*}

\subsection{Noise-free Flow Matching Analysis}
In standard flow matching formulation, the model is trained to learn a mapping from noise to target distributions, while using additional conditioning as an extra input. Instead, we proposed a noise-free flow matching where we directly learn mapping from degraded images distribution to target distribution. This definition, offers several practical advantages in the context of iterative restoration. First, by removing the stochastic noise conditioning, the model avoids the accumulation of noise-induced errors across multiple refinement steps, making the RAR procedure substantially more stable, as shown in Figure 7 of the main paper. Second, the clean latent trajectory enables the LQA module to reliably assess intermediate latents at any iteration without being corrupted by injected noise, ensuring that the quality assessment remains consistent throughout the process. Finally, as illustrated in~\figref{fig:flow_matching_curve}, the noise-free variant converges significantly faster, showing smoother optimization dynamics and reduced variance during training. This property makes iterative assessment during training feasible, allowing the model to perform targeted refinement with stable and efficient updates within the RAR loop.

\subsection{Effectiveness of Proposed Modules} 
To assess the effectiveness of the proposed modules, we provide extended ablations on the integration between the IQA and UIR components across both SD1.5 and SD3.5. As shown in~\tabref{tab:abla_supp}, replacing CLIP~\cite{clip} with DepictQA~\cite{depictqa} consistently improves generalization under unknown and composite degradations. Notably, combining DepictQA~\cite{depictqa} with latent-space restoration already yields performance comparable to AutoDIR~\cite{autodir}, which operates in pixel space and relies on an additional NAFNet~\cite{chu2022nafssr} refinement stage. Further replacing text-based conditioning with aligned embeddings leads to clear improvements across most metrics, highlighting the advantage of a tighter and more direct coupling between IQA and UIR.

\subsection{Investigation of Latent-space RAR}
\begin{figure}[t]
  \centering
  \setlength{\tabcolsep}{1pt}
  \footnotesize
    \begin{tabular}{cccc}
        \includegraphics[width=.225\columnwidth]{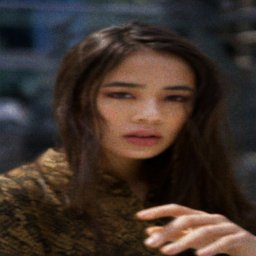} &
        \includegraphics[width=.225\columnwidth]{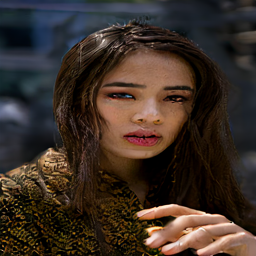} &
        \includegraphics[width=.225\columnwidth]{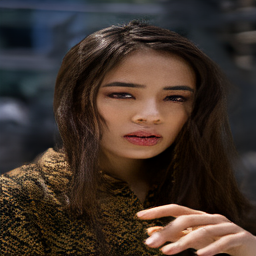} & 
        \includegraphics[width=.225\columnwidth]{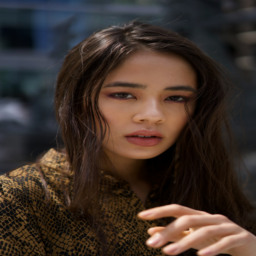} \\
        \includegraphics[width=.225\columnwidth]{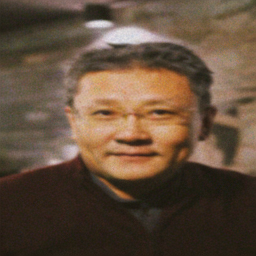} &
        \includegraphics[width=.225\columnwidth]{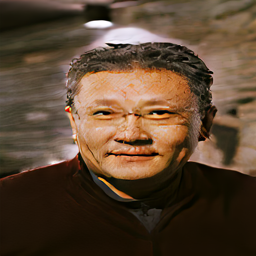} &
        \includegraphics[width=.225\columnwidth]{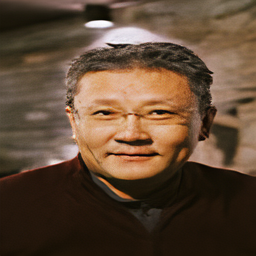} &  
        \includegraphics[width=.225\columnwidth]{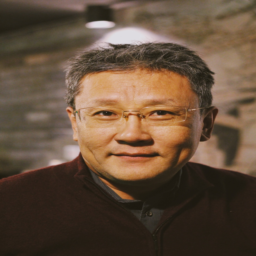} \\
        \includegraphics[width=.225\columnwidth]{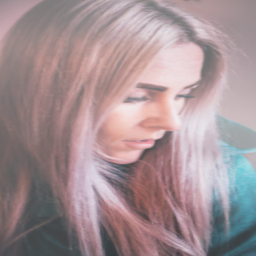} &
        \includegraphics[width=.225\columnwidth]{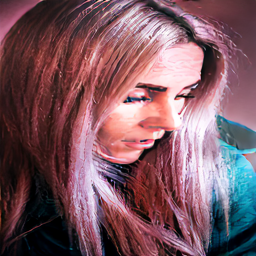} &
        \includegraphics[width=.225\columnwidth]{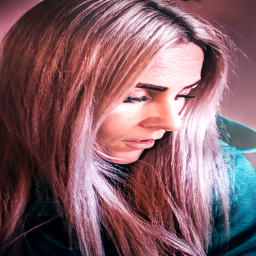} & 
        \includegraphics[width=.225\columnwidth]{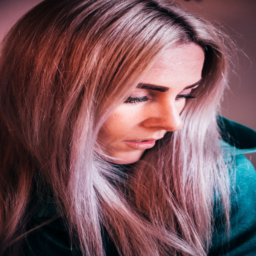} \\
        \includegraphics[width=.225\columnwidth]{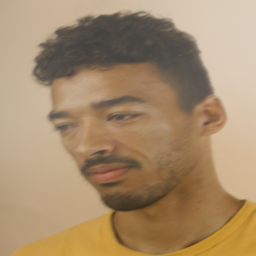} &
        \includegraphics[width=.225\columnwidth]{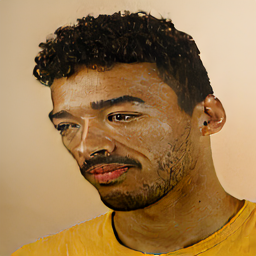} &
        \includegraphics[width=.225\columnwidth]{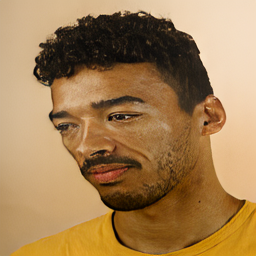} & 
        \includegraphics[width=.225\columnwidth]{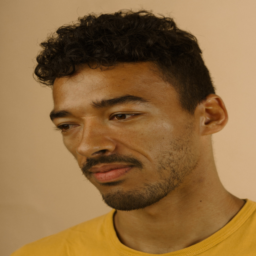} \\
        (a) Input & (b) Pixel & (c) Latent & (d) GT \\
                  &   + Text   & + Embedding &  \\
    \end{tabular}
    \caption{\textbf{Comparison of RAR when iterative operating in pixel space versus latent space.} Please zoom in on the eye region for detailed comparison.}
    \label{fig:abla_text_latent}
    \vspace{-15pt}
\end{figure}

To further analyze the impact of RAR, we compare our latent-space formulation with a vanilla baseline, i.e., "SD3.5 with text conditioning operating in pixel space", which involves three separate stages: (1) generating text-based conditioning through the IQA encoder, and (2) encoding the image again from pixel space using UIR encoder and preforming restoration and (3) repeated VAE encode/decode operations to feedback into the IQA. As shown in Table~3, performing RAR entirely in latent space yields three key advantages.
(1) \textit{Reduced information loss}, as illustrated in~\figref{fig:abla_text_latent}, repeated pixel-space encode/decode operations inevitably wash out fine details (e.g., the eye region), whereas latent refinement preserves structural fidelity across iterations.
(2) \textit{Higher efficiency}, pixel-space pipelines incur substantial overhead from text decoding/encoding (81.19 ms) and VAE (1.22 s), while our latent-space approach requires only a lightweight adapter (14.93 ms) for a 256×256 image on an A100 GPU.
(3) \textit{End-to-end optimization}, the shared latent formulation enables joint optimization of LQA and UIR, resulting in substantial improvements. For example, +13.63\% PSNR and +48.01\% LPIPS on unknown degradations with SD3.5.

\subsection{Quantitative Evaluation for Single Distortion}
In the quantitative evaluation, we further assess the performance of the proposed method across seven single-degradation tasks. 
We note that in single-degradation benchmarks, annotators typically remove only the \textbf{main} distortion (e.g., rain), while other degradations may remain in the image. As our iterative framework removes all identified degradations (see Fig.~5 in main paper), it may deviate from the annotation-specific ground truth, which can lead to reduced PSNR/SSIM. A similar trend is observed in AgenticIR, which is also iterative: it achieves competitive performance on composite degradations but comparatively lower PSNR/SSIM on single-degradation settings. As shown in~\tabref{tab:detail_single_distortion}, this effect is particularly evident in de-raining, where residual degradations are often not addressed by the annotations. Importantly, the PSNR/SSIM drop remains moderate (up to $\simeq6\%$), while our method consistently achieves substantial gains (at least $\simeq 16\%$) across perceptual and no-reference metrics.

\begin{table}[t]
\centering
\caption{Quantitative comparison on \textbf{Single Degradation} tasks. }
\label{tab:detail_single_distortion}
\resizebox{\columnwidth}{!}{%
\begin{tabular}{lcccccc}
\toprule[1.3pt]
\textbf{Method} & \textbf{PSNR} $\uparrow$ & \textbf{SSIM} $\uparrow$ & \textbf{LPIPS} $\downarrow$ & \textbf{CLIP-IQA} $\uparrow$ & \textbf{MUSIQ} $\uparrow$ & \textbf{MANIQA} $\uparrow$ \\
\midrule
\rowcolor{gray!10} \multicolumn{7}{c}{\textbf{Denoise (2 sets)}} \\
\hline
LD        & 22.58 & 0.6250 & 0.1860 & 0.4586 & 44.66 & 0.3481 \\
AirNet    & 29.10 & 0.8030 & 0.2130 & 0.4246 & 47.29 & 0.2997 \\
PromptIR  & 29.89 & 0.8240 & 0.1810 & 0.4124 & 46.22 & 0.2905 \\
AutoDIR   & 29.68 & 0.8320 & 0.1670 & 0.4654 & 43.46 & 0.2996 \\
AgenticIR & 27.89 & 0.8020 & 0.2680 & 0.4653 & 59.30 & 0.3917 \\
\rowcolor{cyan!12} \method{} & 30.21 & 0.8673 & 0.0419 & 0.7144 & 56.73 & 0.4287 \\
\midrule
\rowcolor{gray!10} \multicolumn{7}{c}{\textbf{Derain}} \\
\hline
LD        & 23.21 & 0.6510 & 0.1470 & 0.5505 & 62.31 & 0.3399 \\
AirNet    & 30.99 & 0.9290 & 0.0550 & 0.7338 & 70.50 & 0.4899 \\
PromptIR  & 33.97 & 0.9380 & 0.0490 & 0.7426 & 70.89 & 0.5068 \\
AutoDIR   & 35.09 & 0.9650 & 0.0470 & 0.5992 & 63.87 & 0.3490 \\
AgenticIR & 33.72 & 0.9485 & 0.0713 & 0.5809 & 66.38 & 0.3481 \\
\rowcolor{cyan!12} \method{} & 28.36 & 0.8342 & 0.0565 & 0.6953 & 58.40 & 0.4119 \\
\midrule
\rowcolor{gray!10} \multicolumn{7}{c}{\textbf{Dehaze}} \\
\hline
LD        & 23.49 & 0.7630 & 0.0910 & 0.3633 & 60.88 & 0.3264 \\
AirNet    & 26.52 & 0.9440 & 0.0310 & 0.4960 & 68.41 & 0.4309 \\
PromptIR  & 29.13 & 0.9710 & 0.0220 & 0.5139 & 67.88 & 0.4239 \\
AutoDIR   & 29.34 & 0.9730 & 0.0200 & 0.4004 & 62.54 & 0.3188 \\
AgenticIR & 23.87 & 0.8972 & 0.0834 & 0.3786 & 63.26 & 0.3000 \\
\rowcolor{cyan!12} \method{} & 27.18 & 0.9037 & 0.0332 & 0.4578 & 50.46 & 0.3438 \\
\midrule
\rowcolor{gray!10} \multicolumn{7}{c}{\textbf{Deraindrop}} \\
\hline
LD        & 24.84 & 0.7380 & 0.1040 & 0.4977 & 67.83 & 0.4758 \\
AirNet    & 27.13 & 0.8920 & 0.0940 & 0.4254 & 65.88 & 0.4684 \\
PromptIR  & 27.41 & 0.9000 & 0.0810 & 0.4288 & 66.12 & 0.4740 \\
AutoDIR   & 30.10 & 0.9240 & 0.0630 & 0.4089 & 69.33 & 0.4110 \\
AgenticIR & 21.76 & 0.8320 & 0.2163 & 0.3352 & 61.51 & 0.3149 \\
\rowcolor{cyan!12} \method{} & 25.33 & 0.8061 & 0.0710 & 0.4929 & 55.30 & 0.4155 \\
\midrule
\rowcolor{gray!10} \multicolumn{7}{c}{\textbf{Deblur}} \\
\hline
LD        & 22.53 & 0.6950 & 0.2410 & 0.2170 & 49.64 & 0.2298 \\
AirNet    & 26.50 & 0.8000 & 0.2010 & 0.1921 & 25.14 & 0.1148 \\
PromptIR  & 26.82 & 0.8190 & 0.1980 & 0.2128 & 25.69 & 0.1191 \\
AutoDIR   & 27.07 & 0.8280 & 0.1570 & 0.2494 & 47.30 & 0.2063 \\
AgenticIR & 34.18 & 0.9587 & 0.0411 & 0.2373 & 50.59 & 0.2321 \\
\rowcolor{cyan!12} \method{} & 26.09 & 0.8239 & 0.0700 & 0.3783 & 51.44 & 0.3301 \\
\midrule
\rowcolor{gray!10} \multicolumn{7}{c}{\textbf{Low Light}} \\
\hline
LD        & 18.97 & 0.7700 & 0.1590 & 0.5551 & 71.56 & 0.4883 \\
AirNet    & 21.26 & 0.8180 & 0.2370 & 0.5999 & 38.58 & 0.2848 \\
PromptIR  & 22.42 & 0.8310 & 0.1330 & 0.3062 & 40.44 & 0.2820 \\
AutoDIR   & 22.37 & 0.8880 & 0.1290 & 0.3528 & 67.21 & 0.3422 \\
AgenticIR &  9.89 & 0.4590 & 0.4123 & 0.2973 & 48.69 & 0.2651 \\
\rowcolor{cyan!12} \method{} & 21.75 & 0.8969 & 0.0901 & 0.4179 & 58.55 & 0.4235 \\
\midrule
\rowcolor{gray!10} \multicolumn{7}{c}{\textbf{Super-Resolution}} \\
\hline
LD        & 19.58 & 0.6032 & 0.4266 & 0.3013 & 37.98 & 0.1940 \\
AirNet    & 18.03 & 0.5751 & 0.4429 & 0.3171 & 26.51 & 0.1428 \\
PromptIR  & 19.64 & 0.6286 & 0.3967 & 0.3273 & 26.71 & 0.1480 \\
AutoDIR   & 21.04 & 0.6818 & 0.3148 & 0.3536 & 37.76 & 0.2162 \\
AgenticIR & 19.57 & 0.5580 & 0.5085 & 0.3393 & 41.41 & 0.2322 \\
\rowcolor{cyan!12} \method{} & 22.21 & 0.7326 & 0.1270 & 0.7393 & 61.12 & 0.5338 \\
\bottomrule[1.3pt]
\end{tabular}%
 }
\vspace{-15pt}
\end{table}

\subsection{Investigation of the Stopping Criterion}
We first analyze performance as a function of the number of restoration rounds on both single-degradation tasks and composite degradations (Group C), as shown in~\tabref{tab:round_analysis}. 
Without the stopping mechanism, increasing the number of iterations may introduce mild oversharpening artifacts, which is reflected by gradual degradation in fidelity metrics (e.g., PSNR/SSIM), particularly under extended iteration budgets. 
In contrast, when the stopping rule is enabled, the average number of iterations aligns well with task complexity (e.g., 1.3 for Single and 2.94 for Group C), and performance remains stable.

We further increase the maximum allowed iterations to $N_{\text{max}}=16$ and compare results with and without the stopping rule. As shown in~\tabref{tab:round_number}, without stopping, excessive iterations lead to noticeable degradation in reconstruction quality (e.g., PSNR drops from 19.20 to 17.86 on Group C). However, with the stopping mechanism enabled, both $N_{\text{max}}=4$ and $N_{\text{max}}=16$ produce comparable average iteration counts and nearly identical performance. 
This demonstrates that the termination decision effectively prevents over-restoration and stabilizes the iterative process.

Importantly, the stopping decision is not based on predefined quality thresholds or metric comparisons. Instead, it relies on pairwise image comparison through the LQA model using a fixed prompt template (see~\figref{fig:LQA_tasks}). By leveraging the same LQA module for both degradation assessment and termination decision, the framework maintains consistency between restoration guidance and stopping control, reducing potential divergence during iterative refinement.

\begin{table}[t]
\centering
\caption{\textbf{Stopping Criterion Analysis}. Results for single and composite degradations.}
\label{tab:round_analysis}
\resizebox{\columnwidth}{!}{%
\begin{tabular}{ccccccc}
\toprule[1.3pt]
\textbf{\# rounds} &  \textbf{PSNR} $\uparrow$ & \textbf{SSIM} $\uparrow$ & \textbf{LPIPS} $\downarrow$ & \textbf{CLIPIQA} $\uparrow$ & \textbf{MUSIQ} $\uparrow$ & \textbf{MINIQA} $\uparrow$ \\ 
\midrule
\rowcolor{gray!10} \multicolumn{7}{c}{\textbf{Single Degradation}} \\
\hline
1 & 26.09 & 0.8485 & 0.0672 & 0.5463 & 55.78 & 0.4036 \\
2 & 24.94 & 0.8334 & 0.0760 & 0.5836 & 58.15 & 0.4413 \\ 
3 & 24.32 & 0.8212 & 0.0836 & 0.5970 & 59.31 & 0.4638 \\ 
4 & 23.87 & 0.8127 & 0.0896 & 0.6033 & 59.91 & 0.4774 \\ 
\cellcolor{cyan!12} 1.3 & \cellcolor{cyan!12}25.88 & \cellcolor{cyan!12}0.8378 & \cellcolor{cyan!12}0.0699 & \cellcolor{cyan!12}0.5566 & \cellcolor{cyan!12}56.00 & \cellcolor{cyan!12}0.4125 \\
\midrule
\rowcolor{gray!10} \multicolumn{7}{c}{\textbf{Composite Degradation - Group C}} \\
\hline
1 & 19.47 & 0.6695 & 0.1562 & 0.6176 & 53.43 & 0.4180 \\
2 & 19.36 & 0.6591 & 0.1459 & 0.6592 & 56.92 & 0.4700 \\
3 & 19.28 & 0.6544 & 0.1477 & 0.6650 & 57.84 & 0.4875 \\
4 & 19.20 & 0.6495 & 0.1508 & 0.6683 & 58.33 & 0.4982 \\
\cellcolor{cyan!12} 2.94 & \cellcolor{cyan!12}19.33 & \cellcolor{cyan!12}0.6579 & \cellcolor{cyan!12}0.1489 & \cellcolor{cyan!12}0.6554 & \cellcolor{cyan!12}56.56 & \cellcolor{cyan!12}0.4653 \\ 
\bottomrule[1.3pt]
\end{tabular}%
 }
\end{table}

\begin{table}[t]
\centering
\caption{\textbf{Sensitivity Analysis of Iteration Counts.} Results for composite degradations on Group C.}
\label{tab:round_number}
\resizebox{\columnwidth}{!}{%
\begin{tabular}{ccccccccc}
        \toprule
        \textbf{Stopping} & \textbf{Nmax} & \# \textbf{rounds} &  \textbf{PSNR} $\uparrow$ & \textbf{SSIM} $\uparrow$ & \textbf{LPIPS} $\downarrow$ & \textbf{CLIPIQA} $\uparrow$ & \textbf{MUSIQ} $\uparrow$ & \textbf{MINIQA} $\uparrow$ \\
        \hline
        & 4 & 4        & 19.20 & 0.6495 & 0.1508 & 0.6683 & 58.33 & 0.4982 \\
        \rowcolor{cyan!12} \checkmark & 4 & 2.87     & 19.32 & 0.6578 & 0.1499 & 0.6567 & 56.88 & 0.4694 \\
        \hdashline
        & 16 & 16      & 17.86 & 0.5851 & 0.2243 & 0.6326 & 58.33 & 0.5115 \\
        \rowcolor{cyan!12} \checkmark & 16 & 3.02    & 19.29 & 0.6562 & 0.1519 & 0.6560 & 56.97 & 0.4719 \\
        \bottomrule
        \end{tabular}}
\end{table}

\subsection{Additional Qualitative Evaluations}
\begin{figure*}[t]
  \centering
  \setlength{\tabcolsep}{1pt}
    \footnotesize
    \begin{tabular}{ccccccc}
        \raisebox{0.3\height}{\rotatebox{90}{\small AutoDIR}} & 
        \includegraphics[width=.14\textwidth, height=2.0cm]{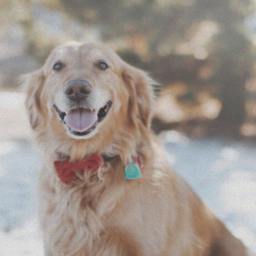} &
        \includegraphics[width=.14\textwidth, height=2.0cm]{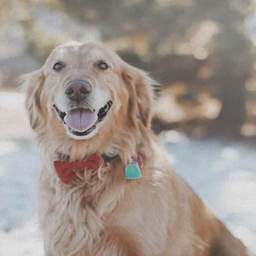} &
        \includegraphics[width=.14\textwidth, height=2.0cm]{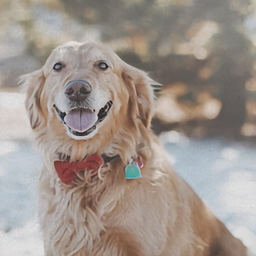} &
        \includegraphics[width=.14\textwidth, height=2.0cm]{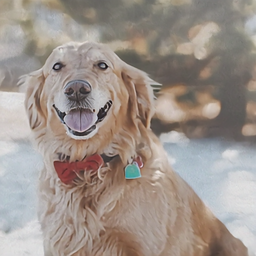} &
        \includegraphics[width=.14\textwidth, height=2.0cm]{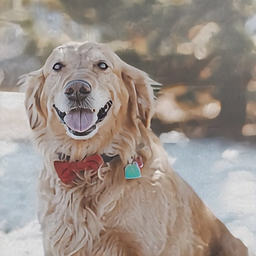} &
        \includegraphics[width=.14\textwidth, height=2.0cm]{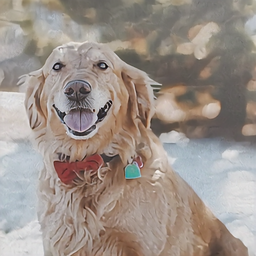} \\
        & Step 1: \textcolor{ForestGreen}{noise} $\rightarrow$ & Step 2: \textcolor{ForestGreen}{blur} $\rightarrow$ & Step 3: \textcolor{ForestGreen}{blur} $\rightarrow$ & Step 4: \textcolor{ForestGreen}{blur} $\rightarrow$ & Step 5: \textcolor{ForestGreen}{blur} $\rightarrow$ & AutoDIR  \\
        \raisebox{0.2\height}{\rotatebox{90}{\small AgenticIR}} & 
        \includegraphics[width=.14\textwidth, height=2.0cm]{figures/Supplementary/Quantative_Comparison/GroupB-FN/Input.png} &
        \includegraphics[width=.14\textwidth, height=2.0cm]{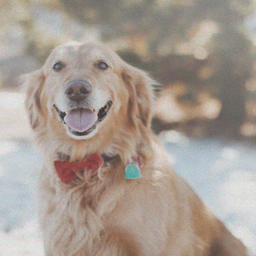} &
        \includegraphics[width=.14\textwidth, height=2.0cm]{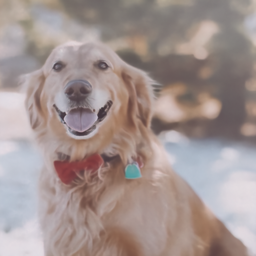} &
        \includegraphics[width=.14\textwidth, height=2.0cm]{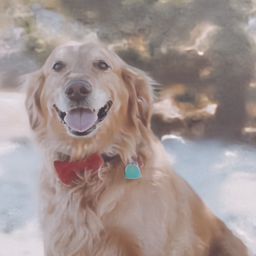} &
        \includegraphics[width=.14\textwidth, height=2.0cm]{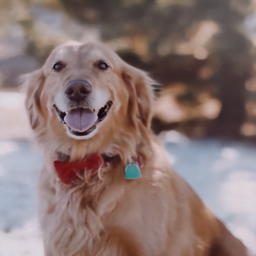} &
        \includegraphics[width=.14\textwidth, height=2.0cm]{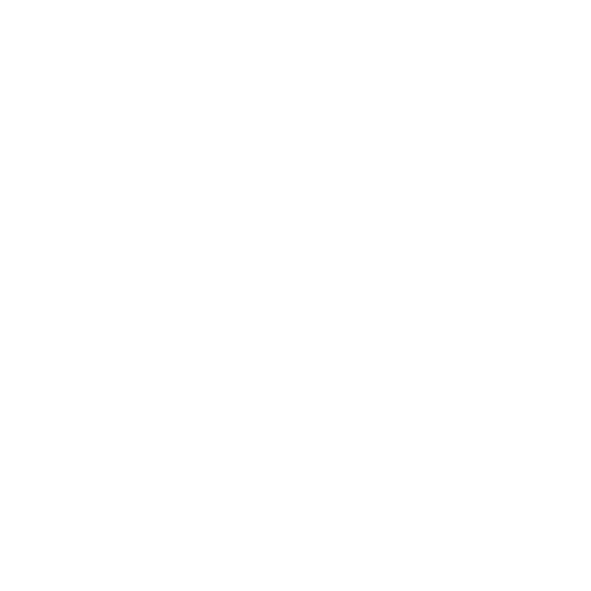} \\
        & Step 1: \textcolor{Fuchsia}{blur (roll)} $\rightarrow$ & Step 2: \textcolor{Fuchsia}{noise} $\rightarrow$ & Step 3: \textcolor{Fuchsia}{blur} $\rightarrow$ & Step 4: \textcolor{Fuchsia}{haze} $\rightarrow$ & AgenticIR &  \\
        \raisebox{\height}{\rotatebox{90}{\small RAR}} & 
        \includegraphics[width=.14\textwidth, height=2.0cm]{figures/Supplementary/Quantative_Comparison/GroupB-FN/Input.png} &
        \includegraphics[width=.14\textwidth, height=2.0cm]{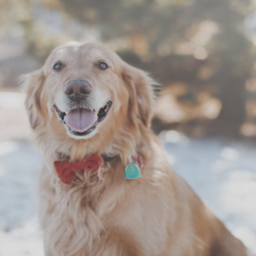} &
        \includegraphics[width=.14\textwidth, height=2.0cm]{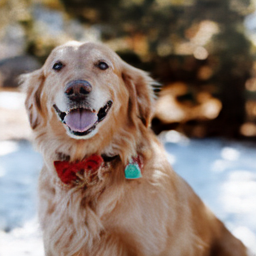} &
        \includegraphics[width=.14\textwidth, height=2.0cm]{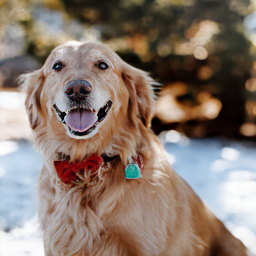} &
        \includegraphics[width=.14\textwidth, height=2.0cm]{figures/Supplementary/Quantative_Comparison/empty.png} &
        \includegraphics[width=.14\textwidth, height=2.0cm]{figures/Supplementary/Quantative_Comparison/empty.png} \\
        & Step 1: \textcolor{blue}{noise} $\rightarrow$ & Step 2: \textcolor{blue}{haze} $\rightarrow$ & Step 3: \textcolor{blue}{blur} $\rightarrow$ & \method{} &  &   \\

        \raisebox{0.3\height}{\rotatebox{90}{\small AutoDIR}} & 
        \includegraphics[width=.14\textwidth, height=2.0cm]{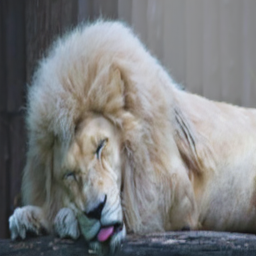} &
        \includegraphics[width=.14\textwidth, height=2.0cm]{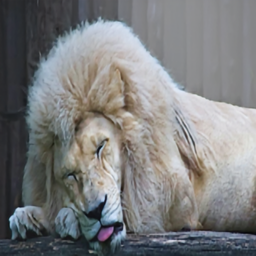} &
        \includegraphics[width=.14\textwidth, height=2.0cm]{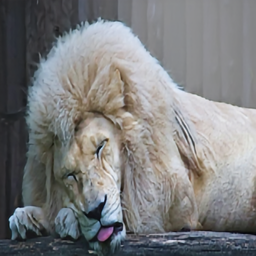} &
        \includegraphics[width=.14\textwidth, height=2.0cm]{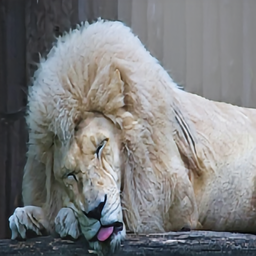} &
        \includegraphics[width=.14\textwidth, height=2.0cm]{figures/Supplementary/Quantative_Comparison/empty.png} &
        \includegraphics[width=.14\textwidth, height=2.0cm]{figures/Supplementary/Quantative_Comparison/empty.png} \\
        & Step 1: \textcolor{ForestGreen}{haze} $\rightarrow$ & Step 2: \textcolor{ForestGreen}{blur} $\rightarrow$ & Step 3: \textcolor{ForestGreen}{resolution} $\rightarrow$ & AutoDIR &  &  \\
        \raisebox{0.2\height}{\rotatebox{90}{\small AgenticIR}} & 
        \includegraphics[width=.14\textwidth, height=2.0cm]{figures/Supplementary/Quantative_Comparison/GroupC-Fog-Blur-LR/Input.png} &
        \includegraphics[width=.14\textwidth, height=2.0cm]{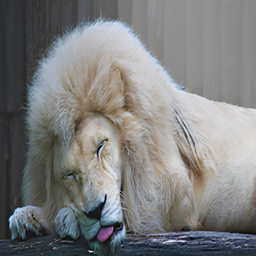} &
        \includegraphics[width=.14\textwidth, height=2.0cm]{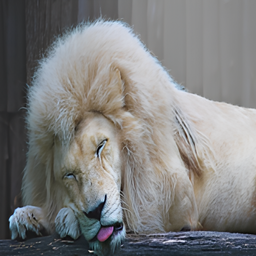} &
        \includegraphics[width=.14\textwidth, height=2.0cm]{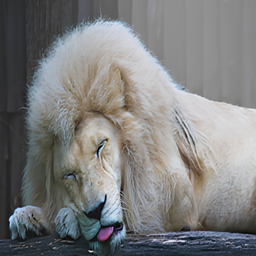} &
        \includegraphics[width=.14\textwidth, height=2.0cm]{figures/Supplementary/Quantative_Comparison/empty.png} &
        \includegraphics[width=.14\textwidth, height=2.0cm]{figures/Supplementary/Quantative_Comparison/empty.png} \\
        & Step 1: \textcolor{Fuchsia}{resolution} $\rightarrow$ & Step 2: \textcolor{Fuchsia}{blur} $\rightarrow$ & Step 3: \textcolor{Fuchsia}{resolution} $\rightarrow$ & AgenticIR &  &  \\
        \raisebox{\height}{\rotatebox{90}{\small RAR}} & 
        \includegraphics[width=.14\textwidth, height=2.0cm]{figures/Supplementary/Quantative_Comparison/GroupC-Fog-Blur-LR/Input.png} &
        \includegraphics[width=.14\textwidth, height=2.0cm]{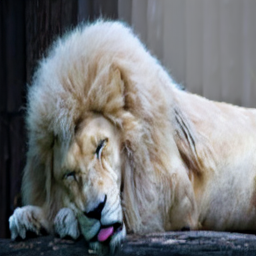} &
        \includegraphics[width=.14\textwidth, height=2.0cm]{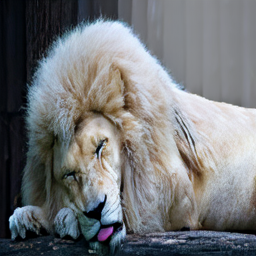} &
        \includegraphics[width=.14\textwidth, height=2.0cm]{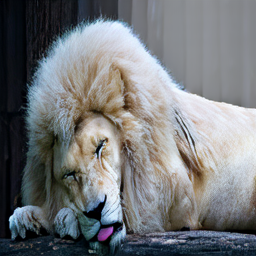} &
        \includegraphics[width=.14\textwidth, height=2.0cm]{figures/Supplementary/Quantative_Comparison/empty.png} &
        \includegraphics[width=.14\textwidth, height=2.0cm]{figures/Supplementary/Quantative_Comparison/empty.png} \\
        & Step 1: \textcolor{blue}{haze} $\rightarrow$ & Step 2: \textcolor{blue}{blur} $\rightarrow$ & Step 3: \textcolor{blue}{resolution} $\rightarrow$ & \method{} &  &   \\

        \raisebox{0.3\height}{\rotatebox{90}{\small AutoDIR}} & 
        \includegraphics[width=.14\textwidth, height=2.0cm]{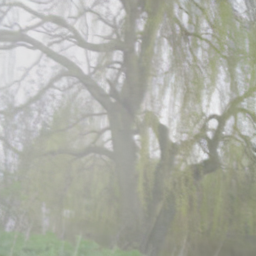} &
        \includegraphics[width=.14\textwidth, height=2.0cm]{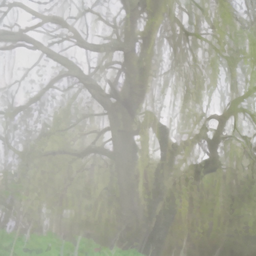} &
        \includegraphics[width=.14\textwidth, height=2.0cm]{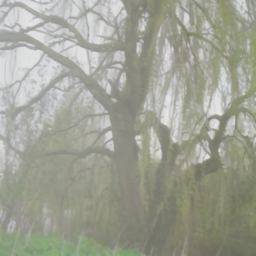} &
        \includegraphics[width=.14\textwidth, height=2.0cm]{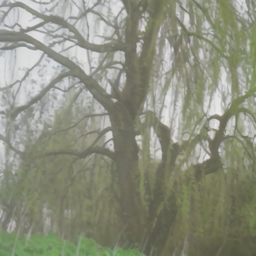} &
        \includegraphics[width=.14\textwidth, height=2.0cm]{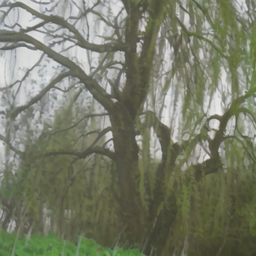} &
        \includegraphics[width=.14\textwidth, height=2.0cm]{figures/Supplementary/Quantative_Comparison/empty.png} \\
        & Step 1: \textcolor{ForestGreen}{blur} $\rightarrow$ & Step 2: \textcolor{ForestGreen}{raindrop} $\rightarrow$ & Step 3: \textcolor{ForestGreen}{haze} $\rightarrow$ & Step 4: \textcolor{ForestGreen}{haze} $\rightarrow$ & AutoDIR &  \\
        \raisebox{0.2\height}{\rotatebox{90}{\small AgenticIR}} & 
        \includegraphics[width=.14\textwidth, height=2.0cm]{figures/Supplementary/Quantative_Comparison/GroupC-Fog-Blur-LR2/Input.png} &
        \includegraphics[width=.14\textwidth, height=2.0cm]{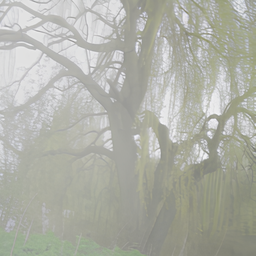} &
        \includegraphics[width=.14\textwidth, height=2.0cm]{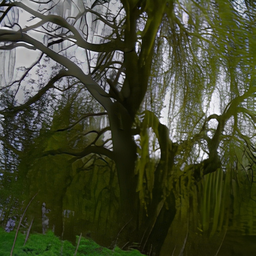} &
        \includegraphics[width=.14\textwidth, height=2.0cm]{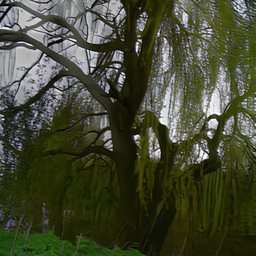} &
        \includegraphics[width=.14\textwidth, height=2.0cm]{figures/Supplementary/Quantative_Comparison/empty.png} &
        \includegraphics[width=.14\textwidth, height=2.0cm]{figures/Supplementary/Quantative_Comparison/empty.png} \\
        & Step 1: \textcolor{Fuchsia}{resolution} $\rightarrow$ & Step 2: \textcolor{Fuchsia}{haze} $\rightarrow$ & Step 3: \textcolor{Fuchsia}{blur} $\rightarrow$ & AgenticIR &  &  \\
        \raisebox{\height}{\rotatebox{90}{\small RAR}} & 
        \includegraphics[width=.14\textwidth, height=2.0cm]{figures/Supplementary/Quantative_Comparison/GroupC-Fog-Blur-LR2/Input.png} &
        \includegraphics[width=.14\textwidth, height=2.0cm]{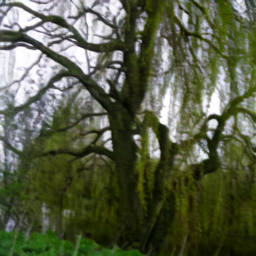} &
        \includegraphics[width=.14\textwidth, height=2.0cm]{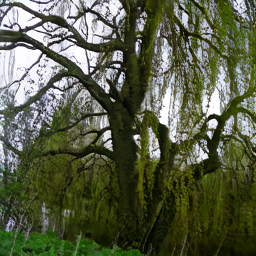} &
        \includegraphics[width=.14\textwidth, height=2.0cm]{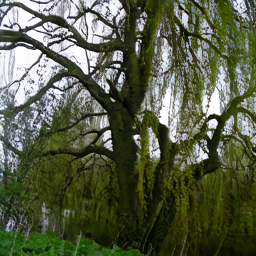} &
        \includegraphics[width=.14\textwidth, height=2.0cm]{figures/Supplementary/Quantative_Comparison/empty.png} &
        \includegraphics[width=.14\textwidth, height=2.0cm]{figures/Supplementary/Quantative_Comparison/empty.png} \\
        & Step 1: \textcolor{blue}{haze} $\rightarrow$ & Step 2: \textcolor{blue}{blur} $\rightarrow$ & Step 3: \textcolor{blue}{resolution} $\rightarrow$ & \method{} &  &   \\

    \end{tabular}
    \caption{\textbf{Qualitative analysis of iterative restoration methods on composite degradations}, including haze, blur, noise, low-resolution conditions.}
    \label{fig:supp_vis1}
\end{figure*}

\begin{figure*}[t]
  \centering
  \setlength{\tabcolsep}{1pt}
    \footnotesize
    \begin{tabular}{ccccccc}
        \raisebox{0.3\height}{\rotatebox{90}{\small AutoDIR}} & 
        \includegraphics[width=.14\textwidth, height=2.0cm]{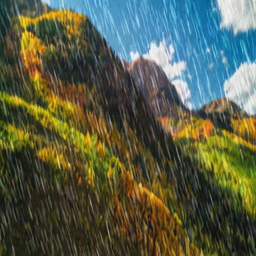} &
        \includegraphics[width=.14\textwidth, height=2.0cm]{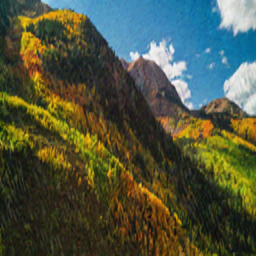} &
        \includegraphics[width=.14\textwidth, height=2.0cm]{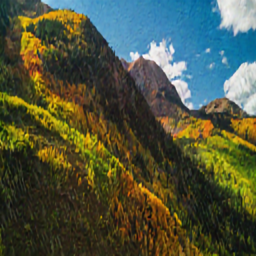} &
        \includegraphics[width=.14\textwidth, height=2.0cm]{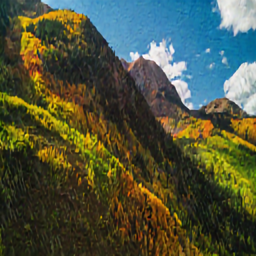} &
        \includegraphics[width=.14\textwidth, height=2.0cm]{figures/Supplementary/Quantative_Comparison/empty.png} &
        \includegraphics[width=.14\textwidth, height=2.0cm]{figures/Supplementary/Quantative_Comparison/empty.png} \\
        & Step 1: \textcolor{ForestGreen}{rain} $\rightarrow$ & Step 2: \textcolor{ForestGreen}{blur} $\rightarrow$ & Step 3: \textcolor{ForestGreen}{blur} $\rightarrow$ & AutoDIR &  &  \\
        \raisebox{0.2\height}{\rotatebox{90}{\small AgenticIR}} & 
        \includegraphics[width=.14\textwidth, height=2.0cm]{figures/Supplementary/Quantative_Comparison/GroupC-Rain-Noise-LR_32/Input.png} &
        \includegraphics[width=.14\textwidth, height=2.0cm]{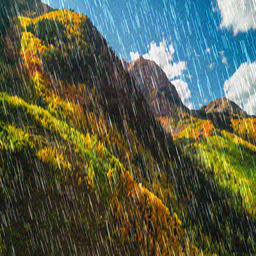} &
        \includegraphics[width=.14\textwidth, height=2.0cm]{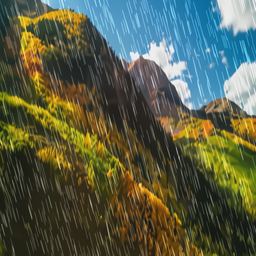} &
        \includegraphics[width=.14\textwidth, height=2.0cm]{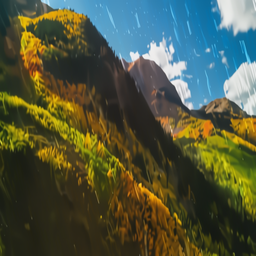} &
        \includegraphics[width=.14\textwidth, height=2.0cm]{figures/Supplementary/Quantative_Comparison/empty.png} &
        \includegraphics[width=.14\textwidth, height=2.0cm]{figures/Supplementary/Quantative_Comparison/empty.png} \\
        & Step 1: \textcolor{Fuchsia}{resolution} $\rightarrow$ & Step 2: \textcolor{Fuchsia}{noise} $\rightarrow$ & Step 3: \textcolor{Fuchsia}{rain} $\rightarrow$ & AgenticIR &  &  \\
        \raisebox{\height}{\rotatebox{90}{\small RAR}} & 
        \includegraphics[width=.14\textwidth, height=2.0cm]{figures/Supplementary/Quantative_Comparison/GroupC-Rain-Noise-LR_32/Input.png} &
        \includegraphics[width=.14\textwidth, height=2.0cm]{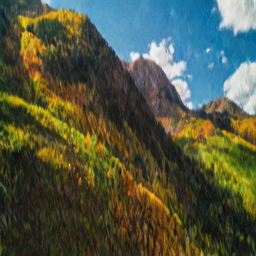} &
        \includegraphics[width=.14\textwidth, height=2.0cm]{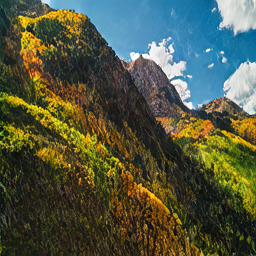} &
        \includegraphics[width=.14\textwidth, height=2.0cm]{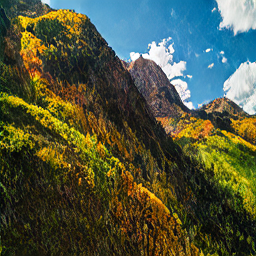} &
        \includegraphics[width=.14\textwidth, height=2.0cm]{figures/Supplementary/Quantative_Comparison/empty.png} &
        \includegraphics[width=.14\textwidth, height=2.0cm]{figures/Supplementary/Quantative_Comparison/empty.png} \\
        & Step 1: \textcolor{blue}{rain} $\rightarrow$ & Step 2: \textcolor{blue}{resolution} $\rightarrow$ & Step 3: \textcolor{blue}{haze} $\rightarrow$ & \method{} &  &   \\

        \raisebox{0.3\height}{\rotatebox{90}{\small AutoDIR}} & 
        \includegraphics[width=.14\textwidth, height=2.0cm]{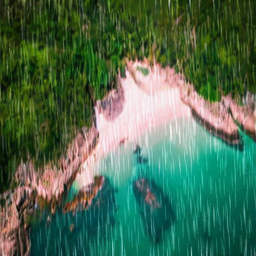} &
        \includegraphics[width=.14\textwidth, height=2.0cm]{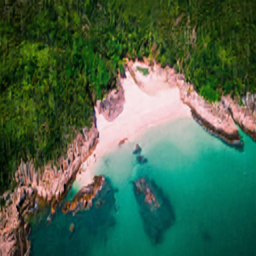} &
        \includegraphics[width=.14\textwidth, height=2.0cm]{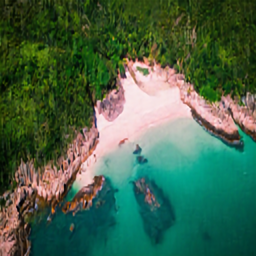} &
        \includegraphics[width=.14\textwidth, height=2.0cm]{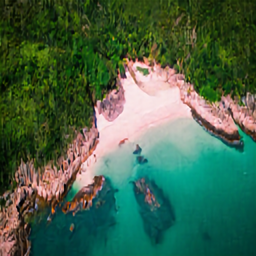} &
        \includegraphics[width=.14\textwidth, height=2.0cm]{figures/Supplementary/Quantative_Comparison/empty.png} &
        \includegraphics[width=.14\textwidth, height=2.0cm]{figures/Supplementary/Quantative_Comparison/empty.png} \\
        & Step 1: \textcolor{ForestGreen}{rain} $\rightarrow$ & Step 2: \textcolor{ForestGreen}{blur} $\rightarrow$ & Step 3: \textcolor{ForestGreen}{blur} $\rightarrow$ & AutoDIR &  &  \\
        \raisebox{0.2\height}{\rotatebox{90}{\small AgenticIR}} & 
        \includegraphics[width=.14\textwidth, height=2.0cm]{figures/Supplementary/Quantative_Comparison/GroupC-Rain-Noise-LR_38/Input.png} &
        \includegraphics[width=.14\textwidth, height=2.0cm]{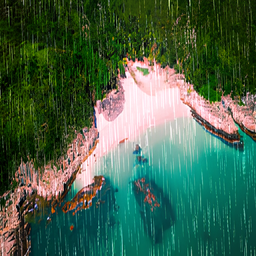} &
        \includegraphics[width=.14\textwidth, height=2.0cm]{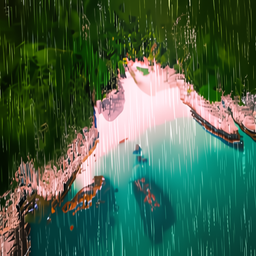} &
        \includegraphics[width=.14\textwidth, height=2.0cm]{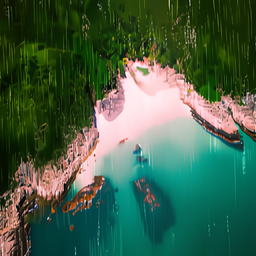} &
        \includegraphics[width=.14\textwidth, height=2.0cm]{figures/Supplementary/Quantative_Comparison/empty.png} &
        \includegraphics[width=.14\textwidth, height=2.0cm]{figures/Supplementary/Quantative_Comparison/empty.png} \\
        & Step 1: \textcolor{Fuchsia}{resolution} $\rightarrow$ & Step 2: \textcolor{Fuchsia}{noise} $\rightarrow$ & Step 3: \textcolor{Fuchsia}{rain} $\rightarrow$ & AgenticIR &  &  \\
        \raisebox{\height}{\rotatebox{90}{\small RAR}} & 
        \includegraphics[width=.14\textwidth, height=2.0cm]{figures/Supplementary/Quantative_Comparison/GroupC-Rain-Noise-LR_38/Input.png} &
        \includegraphics[width=.14\textwidth, height=2.0cm]{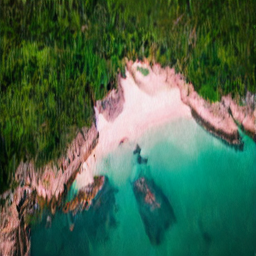} &
        \includegraphics[width=.14\textwidth, height=2.0cm]{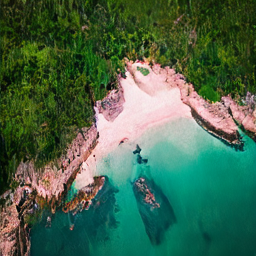} &
        \includegraphics[width=.14\textwidth, height=2.0cm]{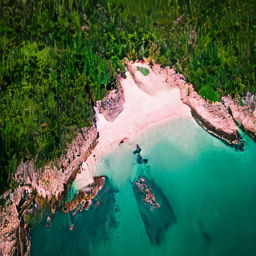} &
        \includegraphics[width=.14\textwidth, height=2.0cm]{figures/Supplementary/Quantative_Comparison/empty.png} &
        \includegraphics[width=.14\textwidth, height=2.0cm]{figures/Supplementary/Quantative_Comparison/empty.png} \\
        & Step 1: \textcolor{blue}{rain} $\rightarrow$ & Step 2: \textcolor{blue}{blur} $\rightarrow$ & Step 3: \textcolor{blue}{resolution} $\rightarrow$ & \method{} &  &   \\

        \raisebox{0.3\height}{\rotatebox{90}{\small AutoDIR}} & 
        \includegraphics[width=.14\textwidth, height=2.0cm]{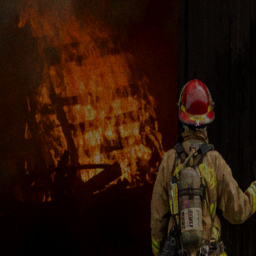} &
        \includegraphics[width=.14\textwidth, height=2.0cm]{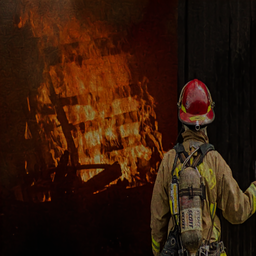} &
        \includegraphics[width=.14\textwidth, height=2.0cm]{figures/Supplementary/Quantative_Comparison/empty.png} &
        \includegraphics[width=.14\textwidth, height=2.0cm]{figures/Supplementary/Quantative_Comparison/empty.png} &
        \includegraphics[width=.14\textwidth, height=2.0cm]{figures/Supplementary/Quantative_Comparison/empty.png} &
        \includegraphics[width=.14\textwidth, height=2.0cm]{figures/Supplementary/Quantative_Comparison/empty.png} \\
        & Step 1: \textcolor{ForestGreen}{noise} $\rightarrow$ & AutoDIR &  &  &  &  \\
        \raisebox{0.2\height}{\rotatebox{90}{\small AgenticIR}} & 
        \includegraphics[width=.14\textwidth, height=2.0cm]{figures/Supplementary/Quantative_Comparison/GroupA-LL-Noise/Input.png} &
        \includegraphics[width=.14\textwidth, height=2.0cm]{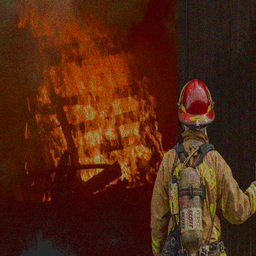} &
        \includegraphics[width=.14\textwidth, height=2.0cm]{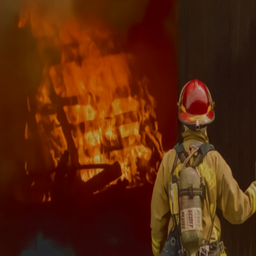} &
        \includegraphics[width=.14\textwidth, height=2.0cm]{figures/Supplementary/Quantative_Comparison/empty.png} &
        \includegraphics[width=.14\textwidth, height=2.0cm]{figures/Supplementary/Quantative_Comparison/empty.png} &
        \includegraphics[width=.14\textwidth, height=2.0cm]{figures/Supplementary/Quantative_Comparison/empty.png} \\
        & Step 1: \textcolor{Fuchsia}{low-light} $\rightarrow$ & Step 2: \textcolor{Fuchsia}{noise} $\rightarrow$ & AgenticIR & & &  \\
        \raisebox{\height}{\rotatebox{90}{\small RAR}} & 
        \includegraphics[width=.14\textwidth, height=2.0cm]{figures/Supplementary/Quantative_Comparison/GroupA-LL-Noise/Input.png} &
        \includegraphics[width=.14\textwidth, height=2.0cm]{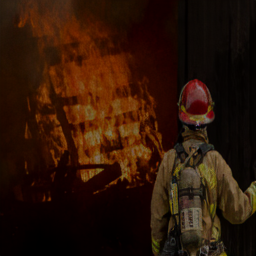} &
        \includegraphics[width=.14\textwidth, height=2.0cm]{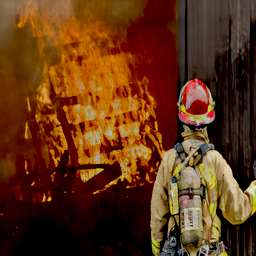} &
        \includegraphics[width=.14\textwidth, height=2.0cm]{figures/Supplementary/Quantative_Comparison/empty.png} &
        \includegraphics[width=.14\textwidth, height=2.0cm]{figures/Supplementary/Quantative_Comparison/empty.png} &
        \includegraphics[width=.14\textwidth, height=2.0cm]{figures/Supplementary/Quantative_Comparison/empty.png} \\
        & Step 1: \textcolor{blue}{noise} $\rightarrow$ & Step 2: \textcolor{blue}{low-light} $\rightarrow$ & \method{} &  &  &  \\
        
    \end{tabular}
    \caption{\textbf{Qualitative analysis of iterative restoration methods on composite degradations,} including rain, noise, low-resolution, low-light.}
    \label{fig:supp_vis2}
\end{figure*}

\begin{figure*}[t]
  \centering
  \setlength{\tabcolsep}{1pt}
    \footnotesize
    \begin{tabular}{cccc|ccc}
        \raisebox{0.3\height}{\rotatebox{90}{\small AutoDIR}} & 
        \includegraphics[width=.14\textwidth, height=2.0cm]{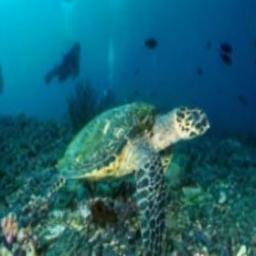} &
        \includegraphics[width=.14\textwidth, height=2.0cm]{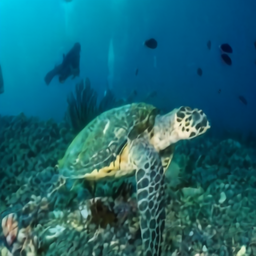} &
        \includegraphics[width=.14\textwidth, height=2.0cm]{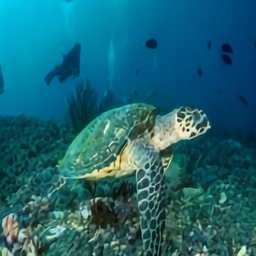} &
        \includegraphics[width=.14\textwidth, height=2.0cm]{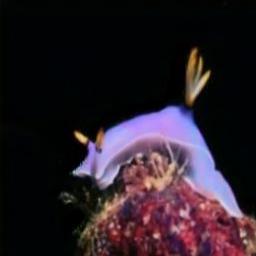} &
        \includegraphics[width=.14\textwidth, height=2.0cm]{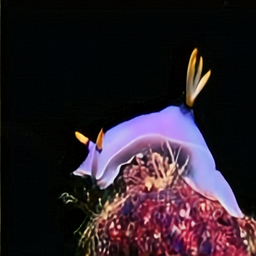} &
        \includegraphics[width=.14\textwidth, height=2.0cm]{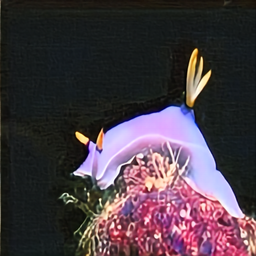} \\
        & Step 1: \textcolor{ForestGreen}{resolution} $\rightarrow$ & Step 2: \textcolor{ForestGreen}{blur} $\rightarrow$ & AutoDIR & Step 1: \textcolor{ForestGreen}{resolution} $\rightarrow$ & Step 2: \textcolor{ForestGreen}{blur} $\rightarrow$ & AutoDIR \\
        \raisebox{0.2\height}{\rotatebox{90}{\small AgenticIR}} & 
        \includegraphics[width=.14\textwidth, height=2.0cm]{figures/Supplementary/Quantative_Comparison/EUVP_267/input.jpg} &
        \includegraphics[width=.14\textwidth, height=2.0cm]{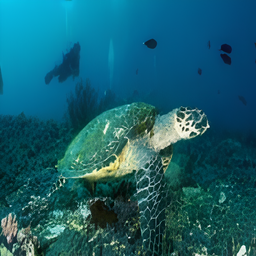} &
        \includegraphics[width=.14\textwidth, height=2.0cm]{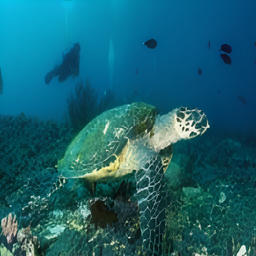} &
        \includegraphics[width=.14\textwidth, height=2.0cm]{figures/Supplementary/Quantative_Comparison/EUVP_415/input.jpg} &
        \includegraphics[width=.14\textwidth, height=2.0cm]{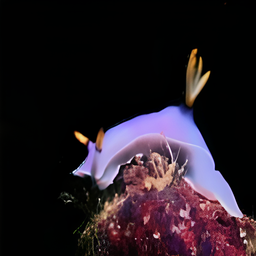} &
        \includegraphics[width=.14\textwidth, height=2.0cm]{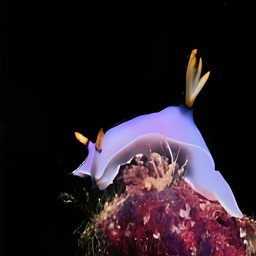} \\
        & Step 1: \textcolor{Fuchsia}{resolution} $\rightarrow$ & Step 2: \textcolor{Fuchsia}{blur} $\rightarrow$ & AgenticIR & Step 1: \textcolor{Fuchsia}{resolution} $\rightarrow$ & Step 2: \textcolor{Fuchsia}{blur} $\rightarrow$ & AgenticIR \\
        \raisebox{\height}{\rotatebox{90}{\small RAR}} & 
        \includegraphics[width=.14\textwidth, height=2.0cm]{figures/Supplementary/Quantative_Comparison/EUVP_267/input.jpg} &
        \includegraphics[width=.14\textwidth, height=2.0cm]{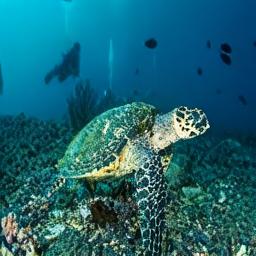} &
        \includegraphics[width=.14\textwidth, height=2.0cm]{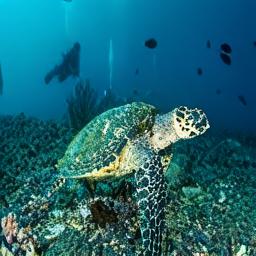} &
        \includegraphics[width=.14\textwidth, height=2.0cm]{figures/Supplementary/Quantative_Comparison/EUVP_415/input.jpg} &
        \includegraphics[width=.14\textwidth, height=2.0cm]{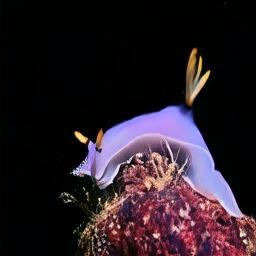} &
        \includegraphics[width=.14\textwidth, height=2.0cm]{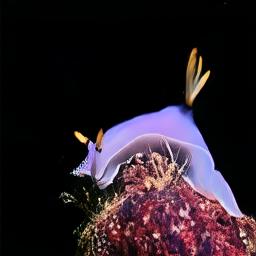} \\
        & Step 1: \textcolor{blue}{blur} $\rightarrow$ & Step 2: \textcolor{blue}{saturation} $\rightarrow$ & \method{} & Step 1: \textcolor{blue}{compression} $\rightarrow$ & Step 2: \textcolor{blue}{contrast} $\rightarrow$ & \method{} \\
    \end{tabular}
    \caption{\textbf{Qualitative analysis of iterative restoration methods on EUVP~\cite{islam2020fast} dataset.}}
    \label{fig:supp_vis3}
\end{figure*}

\begin{figure*}[t]
  \centering
  \setlength{\tabcolsep}{1pt}
    \footnotesize
    \begin{tabular}{cccc|ccc}
        \raisebox{0.3\height}{\rotatebox{90}{\small AutoDIR}} & 
        \includegraphics[width=.14\textwidth, height=2.0cm]{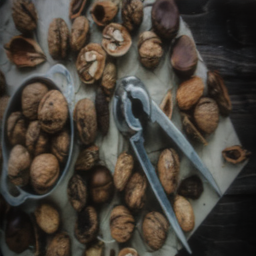} &
        \includegraphics[width=.14\textwidth, height=2.0cm]{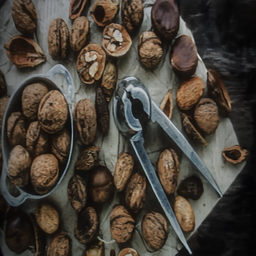} &
        \includegraphics[width=.14\textwidth, height=2.0cm]{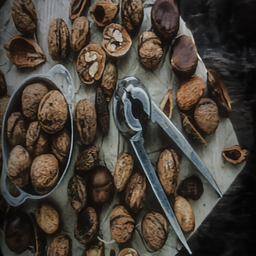} &
        \includegraphics[width=.14\textwidth, height=2.0cm]{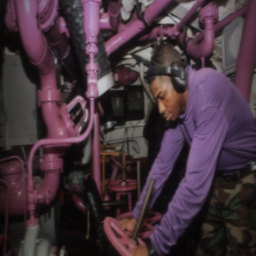} &
        \includegraphics[width=.14\textwidth, height=2.0cm]{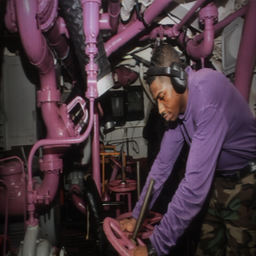} &
        \includegraphics[width=.14\textwidth, height=2.0cm]{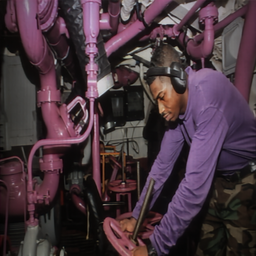} \\
        & Step 1: \textcolor{ForestGreen}{blur} $\rightarrow$ & Step 2: \textcolor{ForestGreen}{blur} $\rightarrow$ & AutoDIR & Step 1: \textcolor{ForestGreen}{blur} $\rightarrow$ & Step 2: \textcolor{ForestGreen}{blur} $\rightarrow$ & AutoDIR \\
        \raisebox{0.2\height}{\rotatebox{90}{\small AgenticIR}} & 
        \includegraphics[width=.14\textwidth, height=2.0cm]{figures/Supplementary/Quantative_Comparison/UDC2/input.png} &
        \includegraphics[width=.14\textwidth, height=2.0cm]{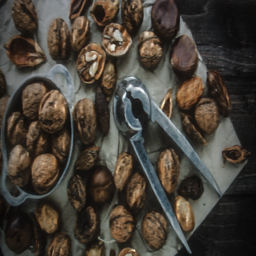} &
        \includegraphics[width=.14\textwidth, height=2.0cm]{figures/Supplementary/Quantative_Comparison/empty.png} &
        \includegraphics[width=.14\textwidth, height=2.0cm]{figures/Supplementary/Quantative_Comparison/UDC18/input.png} &
        \includegraphics[width=.14\textwidth, height=2.0cm]{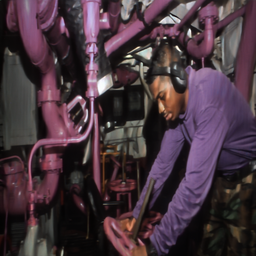} &
        \includegraphics[width=.14\textwidth, height=2.0cm]{figures/Supplementary/Quantative_Comparison/empty.png} \\
        & Step 1: \textcolor{Fuchsia}{blur} $\rightarrow$ & AgenticIR &  & Step 1: \textcolor{Fuchsia}{blur} $\rightarrow$ & AgenticIR & \\
        \raisebox{\height}{\rotatebox{90}{\small RAR}} & 
        \includegraphics[width=.14\textwidth, height=2.0cm]{figures/Supplementary/Quantative_Comparison/UDC2/input.png} &
        \includegraphics[width=.14\textwidth, height=2.0cm]{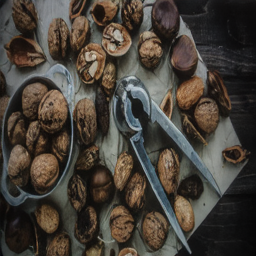} &
        \includegraphics[width=.14\textwidth, height=2.0cm]{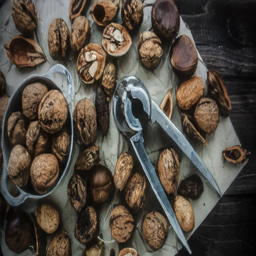} &
        \includegraphics[width=.14\textwidth, height=2.0cm]{figures/Supplementary/Quantative_Comparison/UDC18/input.png} &
        \includegraphics[width=.14\textwidth, height=2.0cm]{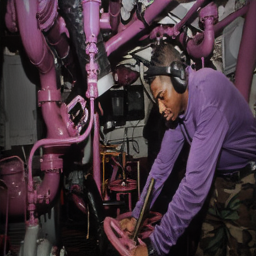} &
        \includegraphics[width=.14\textwidth, height=2.0cm]{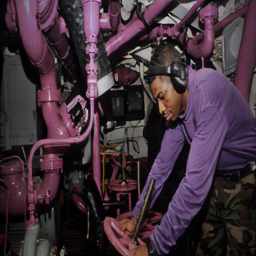} \\
        & Step 1: \textcolor{blue}{blur} $\rightarrow$ & Step 2: \textcolor{blue}{contrast} $\rightarrow$ & \method{} & Step 1: \textcolor{blue}{blur} $\rightarrow$ & Step 2: \textcolor{blue}{contrast} $\rightarrow$ & \method{} \\
    \end{tabular}
    \caption{\textbf{Qualitative analysis of iterative restoration methods on UDC~\cite{zhou2021image} dataset.}}
    \label{fig:supp_vis4}
\end{figure*}

As illustrated in~\figref{fig:supp_vis1} to~\ref{fig:supp_vis4}, our qualitative comparisons reveal clear advantages of \method{} over AutoDIR~\cite{autodir} and AgenticIR~\cite{agenticir} in both efficiency and restoration quality. In terms of efficiency, \method{} consistently completes the iterative restoration process using fewer steps, while AutoDIR~\cite{autodir} frequently generates redundant actions due to its difficulty in reasoning about complex composite degradations, and AgenticIR~\cite{agenticir} incurs substantial computational overhead by invoking multiple task-specific restorers at every step. Regarding restoration quality, AgenticIR~\cite{agenticir} is fundamentally constrained by its single-task restoration model, which often removing textures together with noise, as observed in~\figref{fig:supp_vis1} (Sample 1, Step 2), leading to irreversible detail loss and a blurred final output. In contrast, \method{} leverages a unified image restoration model that jointly models multiple degradation factors within the flow matching, enabling it to more effectively remove heterogeneous degradations while preserving semantic structures and fine textures.

\section{Implementation Details}

\subsection{Training Details}
\method{} is built upon DepictQA~\cite{depictqa} as the LQA backbone and Stable Diffusion 3.5 (SD3.5)~\cite{SD3} as the unified image restoration (UIR) backbone. All experiments are conducted using PyTorch on eight NVIDIA A100 GPUs. The complete training pipeline consists of three primary stages, preceded by an LQA initialization step.
First, we initialize the LQA module with pretrained DepictQA~\cite{depictqa} weights while keeping the IQA head frozen. To align the visual latent representation with the LQA feature space, we replace DepictQA's original image encoder with the UIR image encoder and fine-tune only the image adapter $\mathcal{A}_I$, which maps UIR latents into the IQA feature domain. After the adapter converges, we unfreeze the IQA head with LoRA for a brief joint refinement stage. This step is optimized using Adam with a batch size of 256, trained for 20 epochs for latent alignment and 1 additional epoch for IQA refinement. The learning rate is set to $3 \times 10^{-4}$ with a WarmupDecayLR scheduler. 

Then, We train the embedding-conditioning adapter, $\mathcal{A}_Q$, while keeping all text encoders (CLIP-L, CLIP-G, and T5-XXL) frozen. This stabilizes linguistic conditioning and prevents drift from pretrained language priors. The adapter is trained using Adam with a learning rate of $5 \times 10^{-5}$ and a cosine annealing scheduler for 20 epochs.
Finally, we jointly optimize the full RAR framework, including the LQA module, the adapters, and the pre-tuned UIR backbone, using an iterative training process. We set the number of intermediate assessments to 4 steps and update both restoration inputs and LQA predictions at each step. This stage employs the CAMEWrapper optimizer with a cosine annealing schedule, using a learning rate of $2\times10^{-6}$ and a batch size of 256, trained for 5 epochs.

\begin{algorithm}[t!]
\SetAlgoLined
\textbf{Input:}\\
\quad Degraded image: $I_{\text{deg}}$ \\
\quad Encoder--Decoder: $\mathcal{E}(\cdot), \mathcal{D}(\cdot)$ \\
\quad Latent Quality Assessment: $\mathrm{LQA}(\cdot)$ \\
\quad Unified Image Restoration: $\mathrm{UIR}(\cdot)$ \\
\quad Assessment interval: $T$ \\
\quad Maximum iterations: $N_{\max}$ \\[2pt]

\textbf{Initialization:}\\
\quad Encode image: $z^{0}_{\text{deg}} = \mathcal{E}(I_{\text{deg}})$ \\
\quad Initial assessment: $Q^{0} = \mathrm{LQA}_{\text{identify}}(z^{0}_{\text{deg}})$ \\
\quad Store previous latent: $z^{\text{prev}} = z^{0}_{\text{deg}}$ \\
\quad Set iteration counter: $n = 0$ \\[2pt]

\While{$n < N_{\max}$}{
    \emph{Restore phase (1 to $T$):}\\
    \For{$t = 1$ \KwTo $T$}{
        $z^{\,n+1}_{\text{deg}} = \mathrm{UIR}(z^{\,n}_{\text{deg}},\, t,\, Q^{n})$ \\
        $n = n + 1$ \\
    }

    \emph{Assessment phase:}\\
    $d = \mathrm{LQA}_{\text{verify}}(z^{\text{prev}},\, z^{\,n}_{\text{deg}})$ \\[2pt]

    \If{$d = \texttt{STOP}$}{
        \Return $\mathcal{D}(z^{\text{prev}})$ \\
    }
    \Else{
        $Q^{n} = \mathrm{LQA}_{\text{identify}}(z^{\,n}_{\text{deg}})$ \\
        $z^{\text{prev}} = z^{\,n}_{\text{deg}}$ \\
    }
}
\textbf{Output:} Final restored image: $\mathcal{D}(z^{\,n}_{\text{deg}})$
\label{alg:rar}
\caption{RAR Inference Algorithm}
\end{algorithm}

\subsection{Inference Details}
During the inference stage, \method{} performs iterative image restoration through a restore–assess feedback loop, enabling the outputs to be progressively optimized toward perceptual quality as guided by the LQA module. 
In addition, we adopt a LQA-based stopping criterion that terminates the iterative process once the latent quality verified by LQA does not improve over the previous iteration. 
The complete RAR inference pipeline is summarized in Algorithm 1. 
We set the number of restoration steps to $T=4$ and the maximum inference iterations to $N_{\text{max}}=8$ to prevent excessively long refinement loops.
We employ the Flow-Euler sampler with logit-normal weighting as the flow matching scheduler, which stabilizes the refinement trajectory during iterative restoration. 

\clearpage

{\small
\bibliographystyle{ieee_fullname}
\bibliography{egbib}
}

\end{document}